\documentclass[10pt,twocolumn,letterpaper]{article}

\usepackage{cvpr}      
\usepackage{graphicx}
\usepackage{booktabs}
\usepackage{multirow}
\usepackage{makecell}
\usepackage{array}
\usepackage{tabularx}

\usepackage{graphicx} 
\usepackage{float}   
\usepackage{lipsum}

\usepackage{colortbl}
\usepackage{amsmath}

\definecolor{cvprblue}{rgb}{0.21,0.49,0.74}
\usepackage[pagebackref,breaklinks,colorlinks,allcolors=cvprblue]{hyperref}

\title{Region-Constraint In-Context Generation for Instructional Video Editing\thanks{}}

\author{\normalsize Zhongwei Zhang$^{\dag}$, Fuchen Long$^{\S}$, Wei Li$^{\dag}$, Zhaofan Qiu$^{\S}$, Wu Liu$^{\dag}$, Ting Yao$^{\S}$,  and Tao Mei$^{\S}$\\
	$^{\dag}$\normalsize University of Science and Technology of China \quad $^{\S}$\normalsize HiDream.ai Inc. \\
	{\tt\small\ {\tt\small\{zhwzhang, weili2023\}}@mail.ustc.edu.cn}, {\tt\small\{longfuchen, qiuzhaofan\}@hidream.ai} \\
	{\tt\small\ liuwu@live.cn}, {\tt\small\{tiyao, tmei\}@hidream.ai} \\ [0.3em]
    {\url{https://zhw-zhang.github.io/ReCo-page/}}
}

\begin{document}

\twocolumn[{
	\maketitle
    \vspace{-0.25in}
	\begin{figure}[H]
		\hsize=\textwidth
		\centering
		\includegraphics[width=2.0\linewidth]{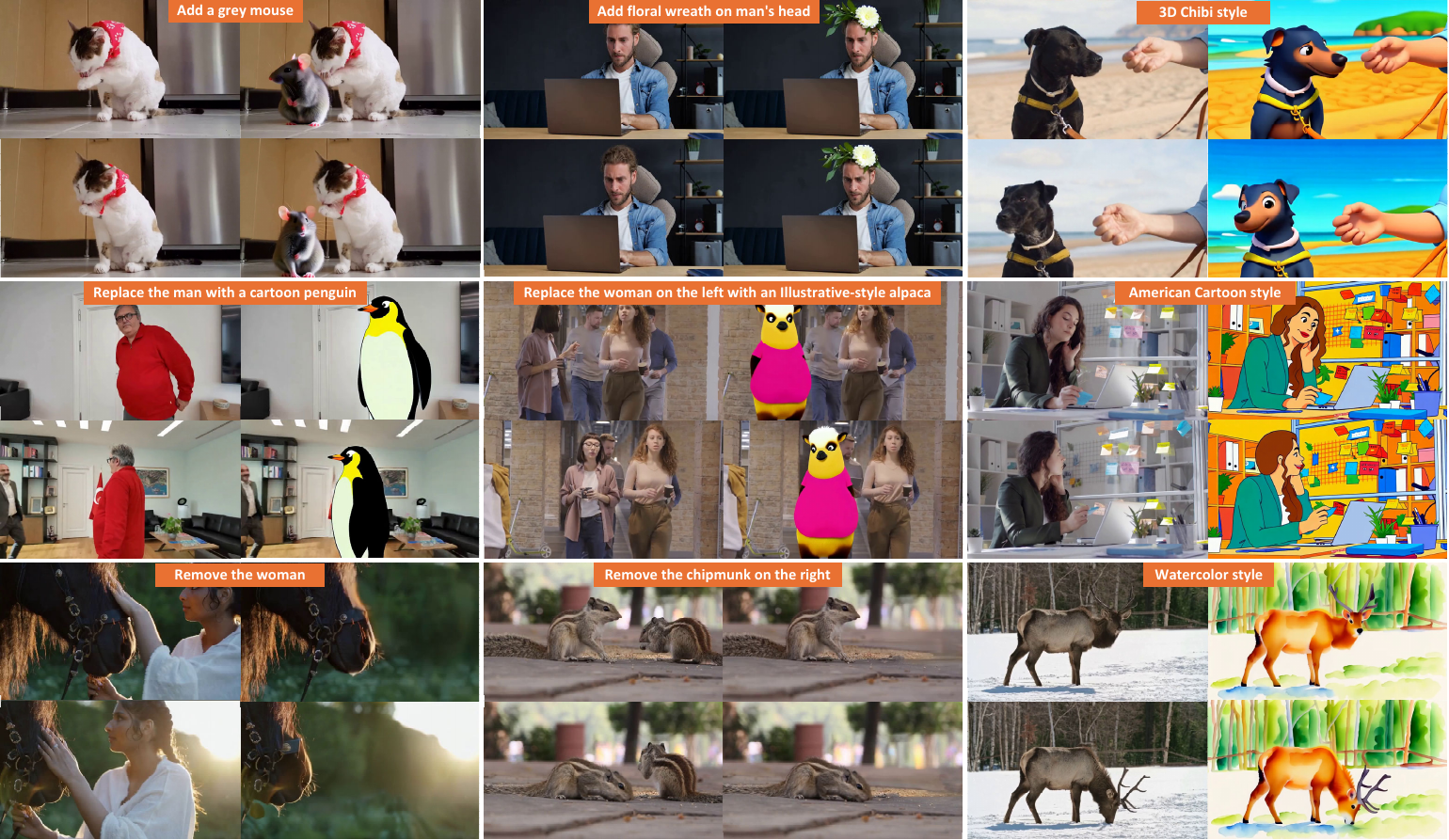}
		\caption{Our ReCo enables video editing based on sole textual instructions, achieving precise and high-fidelity video content modification. ReCo can adeptly handle diverse and challenging video editing tasks, including both local object editing and global style transfer.}
		\label{fig:intro}
	\end{figure}
}]

\renewcommand{\thefootnote}
{\fnsymbol{footnote}}
\footnotetext[1]{This work was performed at HiDream.ai.}

\begin{abstract}
The In-context generation paradigm recently has demonstrated strong power in instructional image editing with both data efficiency and synthesis quality.
Nevertheless, shaping such in-context learning for instruction-based video editing is not trivial.
Without specifying editing regions, the results can suffer from the problem of inaccurate editing regions and the token interference between editing and non-editing areas during denoising. 
To address these, we present ReCo, a new instructional video editing paradigm that novelly delves into constraint modeling between editing and non-editing regions during in-context generation.
Technically, ReCo width-wise concatenates source and target video for joint denoising.
To calibrate video diffusion learning, ReCo capitalizes on two regularization terms, i.e., latent and attention regularization, conducting on one-step backward denoised latents and attention maps, respectively.
The former increases the latent discrepancy of the editing region between source and target videos while reducing that of non-editing areas, emphasizing the modification on editing area and alleviating outside unexpected content generation.  
The latter suppresses the attention of tokens in the editing region to the tokens in counterpart of the source video, thereby mitigating their interference during novel object generation in target video.
Furthermore, we propose a large-scale, high-quality video editing dataset, i.e., ReCo-Data, comprising 500K instruction-video pairs to benefit model training.
Extensive experiments conducted on four major instruction-based video editing tasks demonstrate the superiority of our proposal. 
\end{abstract}





\section{Introduction}
With the rapid advancements in diffusion models \cite{2020ddpm, 2021glide, 2022ldm, 2022photorealistic, 2024SDXL, 2023adding, flowmatchldm2024, flowmatch2022, 2025motionpro, zhang2024trip, liu2025javisdit, sddit,long2024eccv}, instruction-based visual editing for both image and video has garnered significant attention. 
Recent instruction-based image editing models \cite{zhang2025icedit,kontext2025} are capable of editing input images based on natural language instructions without additional condition. Nevertheless, replicating the success attained in image editing within the field of instruction-based video editing is non-trivial. Some promising video editing solutions \cite{jiang2025vace, HunyuanCustom2025} often require input masks to localize editing regions or task-specific configurations, limiting their practicality for use in the real-world.
Steering video editing based on sole textual instruction is still a problem not yet fully explored in the literature.

Inspired by the success of in-context generation paradigm in image editing~\cite{icedit2025,iclora2024} with both data efficiency and generation quality, we construct a joint source-target video diffusion framework for instruction-based video editing.
Due to the inherent temporal complexities, two major challenges are rising when shaping in-context learning for video generation: 1) how to accurately localize the editing region when there is only text instruction? 2) how to further decrease the content interference from source editing region to the novel object generation in target video?
Following the recipe for regional constraint modeling \cite{hu2021region} in visual processing, we address the two issues by modeling the region-wise relationship on both video latents and attention maps. 
We mitigate the first issue through increasing the latent discrepancy in the editing region between source and target videos, and decreasing that in the non-editing areas, which enforces content regeneration in editing area with consistency of background.
To alleviate the second issue, we suppress the attention of tokens in editing region to tokens in the same area of source video, alleviating the token interference from original contents.
This term also encourages the novel object generation to leverage more information from tokens in the background of target video itself, achieving better coherence with background.

By consolidating the idea of region-constraint in-context generation, we present a novel framework dubbed ReCo for instruction-based video editing.
Technically, ReCo first concatenates the source and target videos along the left-right panel, and conducts joint video denoising for editing generation.
In each training step, the paired video latents are first estimated through one-step backward diffusion process.
Then, ReCo calculates latent difference between source and target video latents, and further conducts a pair-wise constraint to increase the latent discrepancy of the editing region and decrease that of non-editing areas.
The similar regularization term is also performed on attention maps of DiT blocks, to suppress the concentration of tokens in the editing region on the tokens of the same region in source video.
Besides, the attention of tokens in the editing region to the background of target video itself are strengthened for harmonious composition between the novel objects and background.
The whole framework is jointly optimized by the flow-matching diffusion loss and the two region-constraint regularization terms.

The main contribution of this work is the new region-constraint in-context generation paradigm for instruction-based video editing. 
Beyond the architecture design, we meticulously construct a large-scale, high-quality video editing dataset, i.e., ReCo-Data, with 500K instruction-video pairs covering a wide spectrum of editing tasks to facilitate community research of instructional video editing.
Extensive experiments further verify the effectiveness of ReCo in terms of both editing accuracy and quality.

\begin{figure*}
\centering
\includegraphics[width=1.0\linewidth]{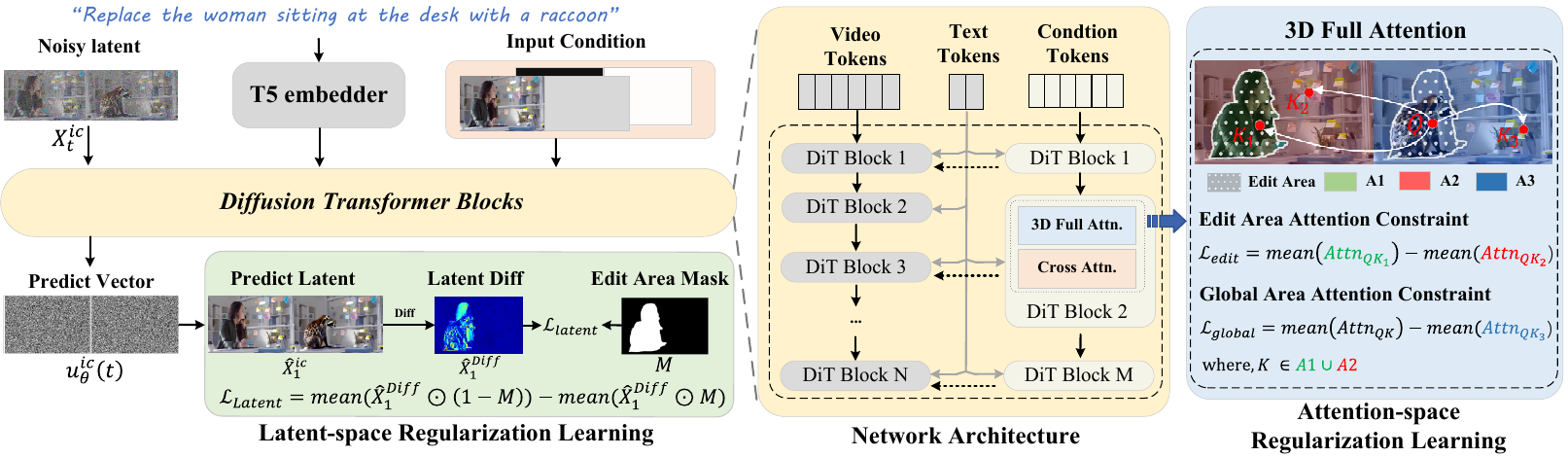}
\vspace{-0.2in}
\caption{
An overview of our ReCo framework. 
We reformulate the instructional video editing task as an \textit{in-context generation} paradigm, guided by the source video and instruction prompt. 
The source video is treated as an explicit condition via feeding it into an auxiliary video condition branch. 
To emphasize editing modifications and alleviate the tokens interference between editing and non-editing areas, ReCo introduces two region-based constraints: 
(1) Latent-space regularization, which increases the latent discrepancy of the editing region between source and target videos while reducing that of non-editing areas. 
(2) Attention-space regularization, which suppresses the attention of the target edit region towards the corresponding region in the source video, thereby mitigating inherent token interference, while simultaneously strengthening the attention on its own generated content.}

\label{fig:framework}
\vspace{-0.2in}
\end{figure*}


\section{Related Work}
\label{sec:related_work}
\textbf{Instruction-based Image Editing.}
Recently, the remarkable progress achieved by text-to-image generation~\cite{2020ddpm,2021glide,2022ldm,2022photorealistic,2024SDXL, 2023adding, flowmatchldm2024,flowmatch2022, 2021glide, 2023instructpix2pix, 2022prompt, FlowEdit, 2024cogvideox,2024vidu,2024hunyuanvideo,wan2025,HaCohen2024LTXVideo} encourages the development of instruction-guided image editing. 
InstructPix2Pix~\cite{2023instructpix2pix}, as one representative work in this domain, establishes a highly effective image editing data construction pipeline and achieves promising editing results. 
Subsequent works treat this data pipeline as a prototype, and refine it to provide more data for training powerful instruction-based editor.
Based on the recipe, multi-modal models like Emu Edit~\cite{2024emuedit}, OmniGen~\cite{ominigen2025}, ICEdit~\cite{icedit2025}, HiDream-E1~\cite{2024hidreami1}, Flux-Kontext~\cite{kontext2025}, Qwen-Image~\cite{qwenimage2025}, and Nano-Banana~\cite{team2023gemini} further unlock the complex capabilities, such as local editing and scene transformations, even without specific fine-tuning.
Nevertheless, it is not a trivial task to replicate the success of image editing in the realm of instructional video editing.
The challenge of video editing lies not only in data scarcity but also in the critical need to simultaneously handle the intricate dependencies between spatial and temporal tokens.
In this work, we address the challenges through an in-context generation paradigm along with regional constraint modeling, supported by our newly constructed ReCo-Data.


\textbf{Instruction-based Video Editing.}
Early attempts~\cite{2023fatezero, rave2024,tokenflow2024,anyv2v2014,videop2p2024,videotome2024,flatten2024,followpose,zichengedit} on instruction-based video editing generally leverage the training-free inference paradigm, which adapt pre-trained text-to-image diffusion models for frame-wise video editing. 
For instance, FateZero~\cite{2023fatezero} edits video frames via DDIM~\cite{2021ddimp} inversion.
However, the lack of temporal modeling leads to the issue of temporal inconsistency. 
To address this, following works such as TokenFlow~\cite{tokenflow2024}, VidToMe~\cite{videotome2024}, FLATTEN~\cite{flatten2024}, and RAVE~\cite{rave2024} employ token-merging or similarity constraints to enhance temporal coherence. 
There are also some approaches (e.g., FlowEdit~\cite{FlowEdit} and FlowDirector~\cite{flowdirector2025}) that exploit advanced text-to-video diffusion models for more accurate diffusion inversion.
Despite the flexibility of the training-free paradigm, the video quality and the model generalization ability are still the inherent limitations.

The primary obstacle for the development of training-based video editing is the profound scarcity of large-scale, high-quality paired training data.
Early approaches overcome this difficulty using one-shot tuning techniques, such as Tune-A-Video~\cite{2023tune} and Video-P2P~\cite{videop2p2024}, but still struggle to achieve ideal editing results.
In another direction, several works~\cite{insv2v2024,genprop2025,insvie2025} push the boundary of video editing in terms of both dataset construction and framework design. 
For example, GenProp~\cite{genprop2025} and Senorita~\cite{senorita2025} formulates the editing as an image propagation mechanism which first edits the first frame and then propagates the content modification into other frames.
More recently, Lucy-Edit~\cite{decart2025lucyedit} and Ditto~\cite{ditto2025} propose to train video editing models directly on source-target videos and text instructions. 
Lucy-Edit concatenates source video latents with denoised video latents as the condition, while Ditto learns the condition through a ControlNet manner. 
Concurrently, the in-context learning~\cite{icedit2025,iclora2024} has been validated in image editing for both data efficiency and generation quality.
In our work, we capitalize on such recipe and further delve into the formulation of region-wise constraint to facilitate accurate video editing.

In summary, our work designs a novel region-constraint in-context generation paradigm for instructional video editing. 
The proposed ReCo contributes by studying not only how to accurately localize editing region, but also how to further reduce the token interference between the editing and non-editing regions for coherent visual modification. 


\section{Our Approach}
Here we will introduce ReCo, a novel region-constrained in-context video generation framework for instructional video editing. 
The overall architecture is illustrated in Figure~\ref{fig:framework}. 
Given a pair of source and target videos, ReCo reformulates the generation process into an in-context learning paradigm, achieved by width-wise concatenating the two videos for joint denoising.
Simultaneously, to ensure the faithful preservation of source video information, we employ an additional video condition branch that explores the condition learning on the source video. 
In the training stage, we introduce two regularization terms, i.e., latent and attention regularization, to benefit accurate video editing learning without pre-specified editing regions.
The latent regularization learning are conducted on the one-step backward denoised latents to amplify region-wise modifications and the consistency of background.
Meanwhile, the attention regularization term suppresses the attention of newly generated objects of the target video on the source video's editing region, thereby decreasing the token interference from the original visual content.


\subsection{Preliminaries: Video DiT Training}
To leverage the prior knowledge from pre-trained video generation models~\cite{2022imagenvideo,2023align,2023make-a-video,sora2024,2024lumiere,2024cogvideox,2024vidu,2024hunyuanvideo,wan2025,HaCohen2024LTXVideo, 2025waver}, we adopt an advanced video diffusion transformer, i.e., Wan-T2V-1.3B~\cite{wan2025}, as the backbone architecture for ReCo. 
To facilitate a clear understanding of our proposal, we first review the training procedure of video DiT.
Typically, most video DiT models are grounded in flow matching~\cite{flowmatch2022,flowmatchldm2024} theory, which provides a theoretically rigorous framework for learning continuous-time generative processes. 
It aims to learn a vector field that smoothly transports samples from a simple prior distribution $P_0$ (e.g., a Gaussian $\mathcal{N}(0, 1)$) to the target data distribution $P_1$.

Given the video latent $x_1$ in training, a random noise sample $x_0 \sim \mathcal{N}(0, 1)$ and a timestep $t \in [0, 1]$ are sampled from a logit-normal distribution. 
Then, $x_0$ is combined with $x_1$ to obtain an intermediate noised latent $x_t$ via the forward diffusion process based on Rectified Flow~\cite{flowmatchldm2024}:
\begin{equation}
x_t = t x_1 + (1-t) x_0 .
\label{eq:xt}
\end{equation}
Then, the ground-truth velocity vector is calculated as:
\begin{equation}
v_t = \frac{dx_t}{dt} = x_1 - x_0 .
\label{eq:vt}
\end{equation}
The video DiT model is learned to estimate this vector via:
\begin{equation}
u_\theta(t) = u(x_t, c, t; \theta) ,
\label{eq:u}
\end{equation}
where $x_t$ is the noisy latent, $\theta$ represents the model parameters, and $c$ is the set of input conditions. For the instructional video editing task, $c$ comprises both the textual instruction and the source video.
Therefore, the training objective $\mathcal{L}$ is defined as the mean squared error (MSE) between the model's output and the ground-truth velocity $v_t$:
\begin{equation}
\mathcal{L} = \mathbb{E}_{x_0, x_1, c, t} \bigl\| u(x_t, c, t; \theta) - v_t \bigr\|^2 .
\label{eq:loss}
\end{equation}
The objective illustrates that the target vector at any given timestep $t$ (i.e., the instantaneous velocity) is simply formulated as $x_1 - x_0$. 
The target is exceptionally clear and stable, making it straightforward for the neural network to learn, which in turn yields high-quality video generation.

\subsection{In-Context Generation for Video Editing}
The in-context generation paradigm has recently demonstrated significant advantages in image editing~\cite{icedit2025,ominicontrol2025,Diptych2025,iclora2024}, particularly in terms of data efficiency and generation quality. Inspired by this, we reformulate the video editing process as in-context generation. 
Technically, given the video latent pair (i.e., the source video $x_1^{src}$ and the target video $x_1^{tar}$), we \emph{width-wise} concatenate them to form a single in-context video latent $x_1^{ic}$ as:
\begin{equation}
x_1^{ic} = [x_1^{src}, x_1^{tar}] .
\end{equation}
During model training, a noise latent $x_0^{ic}$ is sampled from a Gaussian distribution and then added to corrupt $x_1^{ic}$, producing the noisy latent $x_t^{ic}$ which is fed into video DiT for joint source and target video denoising:
\begin{equation}
x_t^{ic} = t x_1^{ic} + (1-t) x_0^{ic} .
\end{equation}
The ground-truth velocity vector is reformulated as:
\begin{equation}
v_t^{ic} = \frac{d x_t^{ic}}{dt} = x_1^{ic} - x_0^{ic} .
\label{eq:vector_ic}
\end{equation}
Consequently, we adapt the training objective Eq.(\ref{eq:loss}) to the in-context generation paradigm and form $\mathcal{L}_{ic}$ as:
\begin{equation}
u_\theta^{ic}(t) = u(x_t^{ic}, c, t; \theta),
\label{eq:u_theta_ic}
\end{equation}
\begin{equation}
\mathcal{L}_{ic} = 
\mathbb{E}_{x_0^{ic}, x_1^{ic}, c, t}
\Bigl\|
u(x_t^{ic}, c, t; \theta) - v_t^{ic}
\Bigr\|^{2} .
\end{equation}

In our in-context generation scenario, the predicted vector $u_\theta^{ic}(t)$ is required to jointly learn both the reconstruction of the source video and the generation of the edited video. 
Due to the high correlation between the source and target videos in editing, such joint learning facilitates strong token interaction, leading to superior video editing performance.
Simultaneously, we employ a video condition branch, as depicted in Figure~\ref{fig:framework}, to ensure that the video condition is comprehensively learned to calibrate video denoising. 
Moreover, we exploit the Low-Rank Adaptation (LoRA) technique for efficient and stable video DiT fine-tuning.


\subsection{Regional Constraint in Latent Space}
The in-context generation benefits token interaction between source and target videos for better instructional video editing.
However, compared to prior advances that require pre-specified edit regions~\cite{jiang2025vace,videopainter2025,HunyuanCustom2025} in video editing, solely relying on textual instruction might still lead to the issue of inaccurate editing region.
To alleviate this limitation, we introduce a regional constraint within the latent space. This mechanism is designed to increase the latent discrepancy of the editing region between source and target videos while reducing that of non-editing areas, amplifying the modification on editing area and alleviating outside unexpected content generation, respectively.


Given the velocity vector $u_\theta^{ic}(t)$ estimated by video DiT with timestep $t$, we first derive the one-step backward denoised latent $\hat{x}_1^{ic}$ based on the Rectified Flow definition as:
\begin{equation}
\hat{x}_1^{ic} = x_t^{ic} + (1-t) u_\theta^{ic}(t) .
\label{eq:backward_denoise}
\end{equation}
The obtained denoised video latent $\hat{x}_1^{ic}$ is then divided along width dimension to get its source and target parts as follows:
\begin{equation}
[\hat{x}_1^{src}, \hat{x}_1^{tar}] = \hat{x}_1^{ic} .
\label{eq:split_latent}
\end{equation}
Next, we calculate the latent difference vector $\hat{X}_1^{Diff}$ between $\hat{x}_1^{src}$ and $\hat{x}_1^{tar}$ through:
\begin{equation}
\hat{X}_1^{Diff} = \left|\hat{x}_1^{tar} - \hat{x}_1^{src}\right| .
\label{eq:latent_diff}
\end{equation}

For a successful editing, we hypothesize that the latent discrepancy should be high within the editing region between the source and target videos, while the non-editing regions should remain unchanged. 
To achieve this, we introduce the latent-space regional constraint $\mathcal{L}_{\text{latent}}$ to regulate DiT training. 
Let $M$ be the binary latent mask indicating the editing region (where $M=1$ denotes regions that should be edited).
$\mathcal{L}_{\text{latent}}$ aims to minimize the mean discrepancy in the non-editing regions while maximizing it within the editing regions, which is computed by:
\begin{equation}
\begin{aligned}
\mathcal{L}_{\text{latent}} = \; & \operatorname{mean}\left( \hat{X}_1^{Diff} \odot (1 - M) \right) \\
& - \operatorname{mean}\left( \hat{X}_1^{Diff} \odot M \right).
\end{aligned}
\label{eq:latent_loss_contrastive}
\end{equation}

\subsection{Regional Constraint in Attention Space}
Besides the region constraint on latent space, the robust learning of in-context generation also necessitates alleviating inherent token interference between the editing and non-editing regions at attention level.
For instance, there should be less concentration on the original contents of editing region in source video, and more attention on its own generated background for better coherence. 
To formulate these relative correlations on attention, we propose to regulate the attention map learning from two perspectives, i.e., the relative relationship within editing region, and the relative relationship within the entire video regions.

As shown in the right part of Figure~\ref{fig:framework}, we first partition the whole area of source-target video pair into three distinct key regions: the source video's editing region (A1), the source video's non-editing region (A2), and the entire target video region (A3).
To formulate the relative relationship within editing region for attention learning, tokens from the target editing region (queries $Q$) should reduce their attention to the corresponding source editing region (keys $K_1$). 
We define this as the \textit{edit attention loss} $\mathcal{L}_{edit}$:
\begin{equation}
\begin{aligned}
\mathcal{L}_{edit} = \operatorname{mean}(Attn_{QK_1}) - \operatorname{mean}(Attn_{QK_2}),
\end{aligned}
\label{eq:attention_edit_loss}
\end{equation}
where $Attn_{QK}$ is the similarity score between queries $Q$ and keys $K$. 
Furthermore, to guarantee coherent integration of generated content with the background, the queries $Q$ should reduce the overall reliance on the entire source video (e.g., keys $K$ in $A1\cup A2$), while focusing more on the contextually relevant of target video regions (e.g., keys $K_3$ in $A3$). 
Therefore, such type of constraint is formulated as the \textit{global attention loss} $\mathcal{L}_{global}$:
\begin{equation}
\begin{aligned}
\mathcal{L}_{global} = \operatorname{mean}(Attn_{QK}) - \operatorname{mean}(Attn_{QK_3}),
\end{aligned}
\label{eq:attention_global_loss}
\end{equation}
The attention-space regional constraint is thus defined as the sum of both two components:
\begin{equation}
\begin{aligned}
\mathcal{L}_{attn} = \mathcal{L}_{edit} + \mathcal{L}_{global}.
\end{aligned}
\label{eq:attn_final}
\end{equation}

Finally, the overall training objective in our ReCo is formulated as a multi-task loss by integrating basic in-context flow matching loss $\mathcal{L}_{ic}$ and two regional constraints in both latent space $\mathcal{L}_{\text{latent}}$ and attention space $\mathcal{L}_{attn}$:
\begin{equation}
\mathcal{L} = \mathcal{L}_{ic} + \lambda_1 \mathcal{L}_{\text{latent}} + \lambda_2 \mathcal{L}_{attn},
\label{eq:total_loss}
\end{equation}
where $\lambda_1$ and $\lambda_2$ are trade-off parameters.
The two constraints emphasize more accurate editing regions and the learning of correct token relationships, mitigating token interference for more natural video content generation.

\begin{table*}
	\centering
	\caption{Performance comparisons on four video editing tasks (i.e., add object, replace object, remove object and style transfer).
		We evaluate the video editing quality by feeding the source and target video pair into Gemini-2.5-Flash-Thinking~\cite{team2023gemini} and asking the VLLM to give the rating from three major perspectives:
		(1) \textit{Edit Accuracy} includes the sub-dimensions of Semantic Accuracy (SA), Scope Precision (SP), and Content Preservation (CP);
		(2) \textit{Video Naturalness} contains Appearance Naturalness (VN), Scale Naturalness (SN), and Motion Naturalness (MN);
		(3) \textit{Video Quality} includes Visual Fidelity (VF), Temporal Stability (TS), and Edit Stability (ES).
		The range of the score for each evaluating sub-dimension is from $0$ to $10$ (higher score is better).
		We also report the per-category scores (i.e., $S_{EA}$, $S_{VN}$, $S_{VQ}$) by computing the geometric mean of all sub-dimension scores of each major perspective, and the overall averaged score $S$. 
		}
    \vspace{-0.09in}
	\label{tab:tasks-methods-final}
	\setlength{\tabcolsep}{0.9em}
	\resizebox{\textwidth}{!}{%
		\begin{tabular}{c c | ccc | ccc | ccc | cccc}
			\toprule
			\multirow{2}{*}{\textbf{Task}} & \multirow{2}{*}{\textbf{Approach}} &
			\multicolumn{3}{c|}{\textbf{Edit Accuracy (EA)}} &
			\multicolumn{3}{c|}{\textbf{Video Naturalness (VN)}} &
			\multicolumn{3}{c|}{\textbf{Video Quality (VQ)}} &
			\multicolumn{4}{c}{\textbf{Average Score}} \\
			\cmidrule{3-15}
			& &
			{SA} & {SP} & {CP} &
			{AN} & {SN} & {MN} &
			{VF} & {TS} & {ES} &
			$S_{EA}$ & $S_{VN}$ & $S_{VQ}$ & $S$ \\

			\midrule
			\multirow{4}{*}{\textbf{Add}} 
			& InsViE~\cite{insvie2025} & 2.60 & 2.79 & 2.78 & 2.33 & 3.98 & 3.74 & 3.71 & 3.91 & 3.58 & 2.60 & 3.10 & 3.46 & 3.05  \\
			& Lucy-Edit~\cite{decart2025lucyedit} & 6.27 & 6.32 & 7.75 & 4.63 & 7.08 & 6.08 & 6.31 & 6.82 & 7.57 & 6.47 & 5.70 & 6.77 & 6.31 \\
			& Ditto~\cite{ditto2025} & 7.46 & 7.24 & 6.30 & 6.30 & \textbf{8.85} & \textbf{8.30} & \textbf{8.13} & 8.55 & 9.03 & 6.70 & \textbf{7.57} & 8.41 & 7.56 \\
			& \cellcolor{gray!20} ReCo & \cellcolor{gray!20} \textbf{8.65} & \cellcolor{gray!20} \textbf{8.40} & \cellcolor{gray!20} \textbf{9.22} & \cellcolor{gray!20} \textbf{6.39} & \cellcolor{gray!20} 8.78 & \cellcolor{gray!20} 8.28 & \cellcolor{gray!20} 8.02 & \cellcolor{gray!20} \textbf{8.61} & \cellcolor{gray!20} \textbf{9.61} & \cellcolor{gray!20} \textbf{8.54} & \cellcolor{gray!20} 7.55 & \cellcolor{gray!20} \textbf{8.61} & \cellcolor{gray!20} \textbf{8.23} \\
			\midrule
			\multirow{4}{*}{\textbf{Replace}} 
			& InsViE~\cite{insvie2025} & 1.89 & 2.38 & 2.48 & 2.58 & 5.25 & 5.05 & 3.76 & 4.00 & 3.52 & 2.10 & 3.91 & 3.49 & 3.17  \\
			& Lucy-Edit~\cite{decart2025lucyedit} & 6.57 & 7.49 & 7.73 & 5.13 & 7.46 & 6.65 & 6.32 & 6.64 & 8.08 & 7.08 & 6.21 & 6.88 & 6.72 \\
			& Ditto~\cite{ditto2025} & 4.95 & 4.83 & 4.79 & 5.81 & 8.63 & 8.10 & 7.55 & 7.95 & 8.71 & 4.56 & 7.21 & 7.96 & 6.58  \\
			& \cellcolor{gray!20} ReCo & \cellcolor{gray!20} \textbf{9.38} & \cellcolor{gray!20} \textbf{9.43} & \cellcolor{gray!20} \textbf{9.59} & \cellcolor{gray!20} \textbf{7.07} & \cellcolor{gray!20} \textbf{8.87} & \cellcolor{gray!20} \textbf{8.47} & \cellcolor{gray!20} \textbf{8.19} & \cellcolor{gray!20} \textbf{8.65} & \cellcolor{gray!20} \textbf{9.67} &
			\cellcolor{gray!20} \textbf{9.43} & \cellcolor{gray!20} \textbf{8.01} & \cellcolor{gray!20} \textbf{8.77} &  \cellcolor{gray!20} \textbf{8.74} \\
			\midrule
			\multirow{3}{*}{\textbf{Remove}}  
			& InsViE~\cite{insvie2025} & 2.53 & 2.49 & 2.44 & 2.63 & 4.87 & 4.72 & 3.41 & 3.67 & 3.40 & 2.44 & 3.76 & 3.29 & 3.16 \\
			& VACE~\cite{jiang2025vace}& 4.58 & 4.58 & 4.56 & 4.96 & 6.09 & 5.89 & 5.48 & 5.50 & 5.57 & 4.57 & 5.43 & 5.56 & 5.19  \\
		    & \cellcolor{gray!20} ReCo & \cellcolor{gray!20} \textbf{7.43} & \cellcolor{gray!20} \textbf{7.43} & \cellcolor{gray!20} \textbf{7.17} & \cellcolor{gray!20} \textbf{6.20} & \cellcolor{gray!20} \textbf{7.43} & \cellcolor{gray!20} \textbf{7.30} & \cellcolor{gray!20} \textbf{6.48} & \cellcolor{gray!20} \textbf{6.63} & \cellcolor{gray!20} \textbf{7.68} & \cellcolor{gray!20} \textbf{7.28} & \cellcolor{gray!20} \textbf{6.90} &  \cellcolor{gray!20} \textbf{6.82} & \cellcolor{gray!20} \textbf{7.00} \\
			\midrule
			\multirow{4}{*}{\textbf{Style}}  
            & InsViE~\cite{insvie2025} & 7.59 & 8.86 & 8.49 & 6.77 & 9.14 & 9.28 & 7.13 & 6.40 & 8.99 & 8.17 & 8.21 & 7.35 & 7.91 \\
		
			& Lucy-Edit~\cite{decart2025lucyedit} & 3.73 & 5.59 & 5.39 & 4.20 & 5.88 & 5.88 & 4.44 & 4.17 & 5.87 & 4.65 & 4.67 & 5.17 & 4.83 \\

			& Ditto~\cite{ditto2025} & 9.10 & 9.36 & 9.26 & 8.25 & 9.51 & 9.58 & 8.33 & 8.33 & 9.77 & 9.20 & 9.07 & 8.77 & 9.01  \\
            
			& \cellcolor{gray!20} ReCo & \cellcolor{gray!20} \textbf{9.11} & \cellcolor{gray!20} \textbf{9.82} & \cellcolor{gray!20} \textbf{9.54} & \cellcolor{gray!20} \textbf{8.43} & \cellcolor{gray!20} \textbf{9.55} & \cellcolor{gray!20} \textbf{9.70} & \cellcolor{gray!20} \textbf{8.61} & \cellcolor{gray!20} \textbf{8.35} & \cellcolor{gray!20} \textbf{9.87} & \cellcolor{gray!20} \textbf{9.42} & \cellcolor{gray!20} \textbf{9.19} & \cellcolor{gray!20} \textbf{8.90} & \cellcolor{gray!20} \textbf{9.17} \\
			\bottomrule
		\end{tabular}
	}
    \vspace{-0.2in}
\end{table*}

\section{Experiments}
\subsection{Experimental Settings}

\begin{figure}
	\centering
\includegraphics[width=0.82\linewidth]{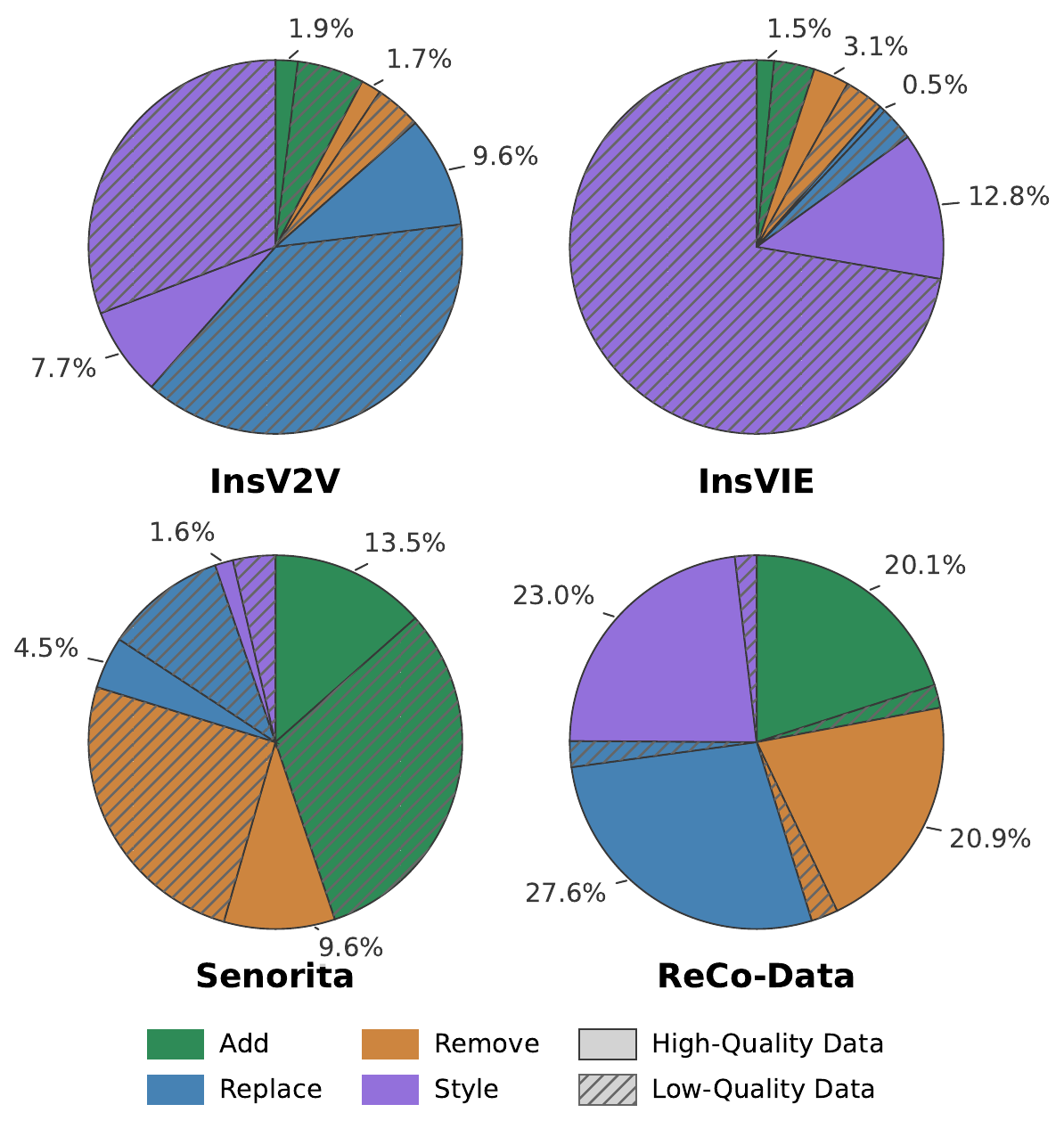}
\vspace{-0.15in}
\caption{Comparison between existing video editing datasets and our ReCo-Data. Ours features the most balanced data distribution and has a higher ratio of the high-quality samples.}
\label{fig:dataset}
\vspace{-0.2in}
\end{figure}

\textbf{Datasets.} 
Despite recent great progress in instructional video editing, a significant bottleneck still remains: the lack of a large-scale, high-quality training dataset. To address this challenge, we introduce the \textbf{ReCo-Data}, which is meticulously curated to support four major video editing tasks: instance-level object adding, removing, and replacing, and the global video stylization.
Our data construction pipeline involves six main stages: (1) raw data pre-process; (2) object segmentation; (3) instruction generation using VLLMs (i.e., Gemini-2.5-Flash-Thinking~\cite{team2023gemini}); (4) condition pairs construction; (5) video synthesis using VACE~\cite{jiang2025vace}; and (6) video filtering and re-captioning with VLLMs. 
More details are provided in the supplementary material.
Ultimately, we construct ReCo-Data with 500K high-quality instruction-video pairs. 
Each video clip contains $81$ frames with the resolution of $480\times 832$.
The video duration is $5.0$ seconds.

We compare ReCo-Data with existing video editing datasets in terms of the ratio of high-quality samples, which reflects the usability and overall quality of the dataset.
Specifically, we randomly sample 200 video editing pairs from each editing task across all datasets, and invite $10$ evaluators to qualitatively assess the video editing quality. 
As shown in Figure~\ref{fig:dataset}, the ratio of high-quality samples in existing datasets (i.e., InsV2V~\cite{insv2v2024}, InsVIE~\cite{insvie2025}, and Senorita~\cite{senorita2025}) is usually low ($17.9\%\sim 29.2\%$).
It indicates that these datasets have not undergone rigorous data cleaning processes, and the large number of low-quality samples makes them suboptimal for training high-performing instructional video editing models.
Besides, the cost of data re-cleaning is extremely high, while the potential benefit is minimal due to the low frame rate, low resolution, and poor synthesis quality of previous datasets.
Instead, our ReCo-Data has a very high proportion ($91.6\%$) of high-quality samples and a well-balanced data distribution across different tasks. 
It can be readily used for model training without any data pre-processing.
The usability of ReCo-Data is also verified by the training of our model.

\textbf{Benchmarks.} 
We construct a video editing evaluation benchmark which contains 480 video-instruction pairs, 120 pairs for each of the four video editing tasks. 
Since traditional metrics usually struggle to accurately and comprehensively evaluate video editing across various dimensions, we follow the image editing advance \cite{ominiedit2025}, and employ a VLLM as the referee for evaluation.
Considering the inherent complexity of video data, we extended the image editing metrics \cite{ominiedit2025} and construct a diverse set of evaluation dimensions tailored for video editing.
We measure the video editing from three main aspects: (1) \textit{Edit Accuracy}, with the sub-dimensions of Semantic Accuracy (SA), Scope Precision (SP), and Content Preservation (CP); (2) \textit{Video Naturalness}, which includes Appearance Naturalness (AN), Scale Naturalness (SN), and Motion Naturalness (MN); and (3) \textit{Video Quality}, with the dimensions of Visual Fidelity (VF), Temporal Stability (TS), and Edit Stability (ES).
We obtain the per-category scores (i.e., $S_{EA}$, $S_{VN}$, $S_{VQ}$) by calculating the geometric mean of all sub-dimension scores of each major perspective.
The overall averaged score ($S$) is the arithmetic mean of the three per-category scores.
Specifically, we feed the source and generated video pair, and the predefined system instructions into Gemini-2.5-Flash-Thinking~\cite{team2023gemini}, and ask it to give the rating for video editing from all the nine sub-dimensions. 
More details about the benchmark construction and the full evaluation protocol are provided in the supplementary material.

\textbf{Implementation Details.}
In ReCo, we employ Wan-T2V-1.3B~\cite{wan2025} as our base architecture. Each training sample is an $81$-frame video clip, with the frame rate of $16$ fps and the resolution of $480 \times 832$. 
For mask generation to align the resolution of video latents, we first encode the editing mask via VAE~\cite{vae} and then apply $k$-means clustering to binarize them.
We set the rank of the LoRA as $128$. 
ReCo is trained using the AdamW optimizer with a two-stage learning rate schedule: the model is first trained with a learning rate of $1 \times 10^{-4}$ to achieve stable convergence, followed by a fine-tuning stage using a lower learning rate of $2 \times 10^{-5}$ for further refinement. 
All experiments are conducted on $24$ NVIDIA A800 GPUs with a mini-batch size of $24$.


\begin{figure*}
	\centering
\includegraphics[width=0.93\linewidth]{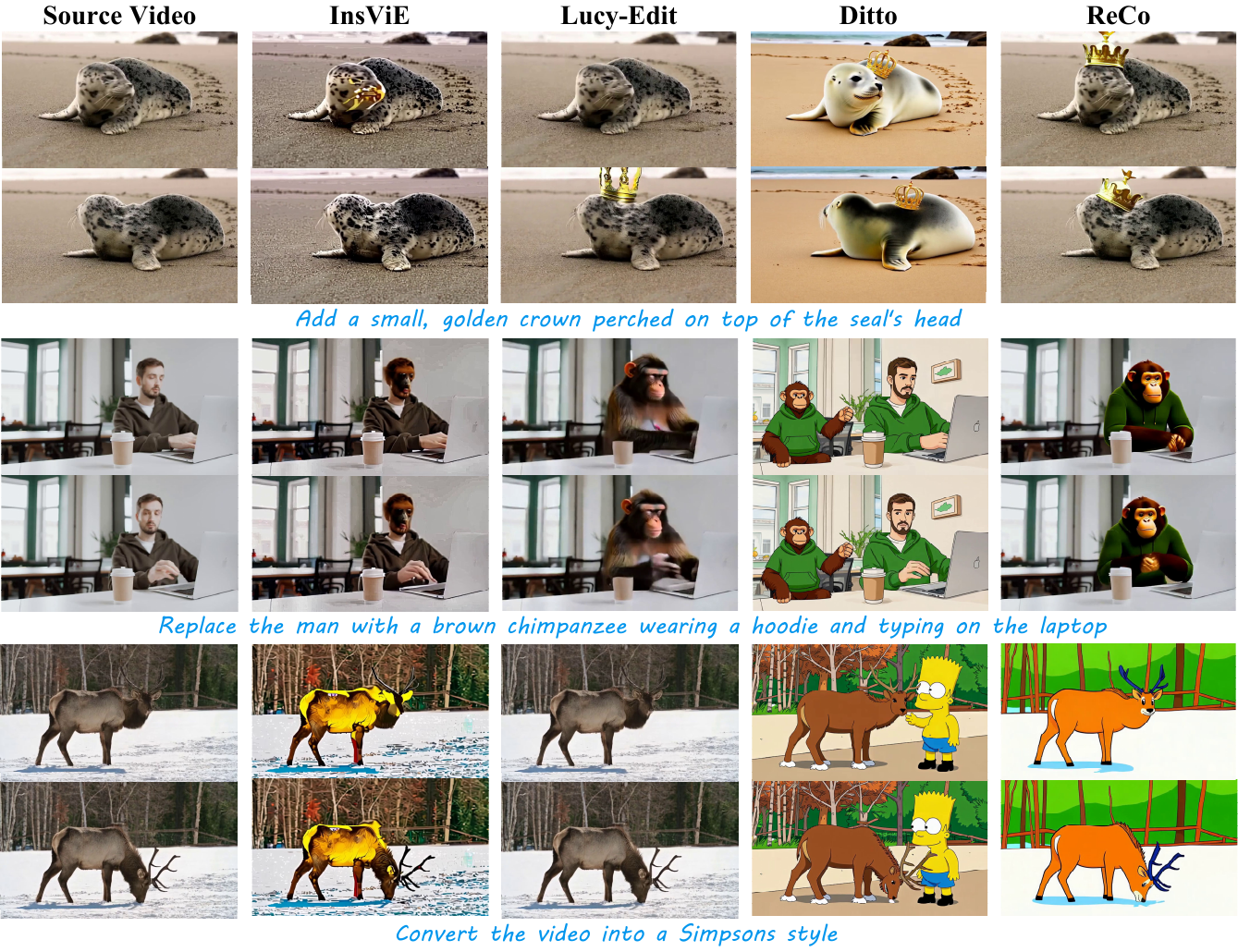}
\vspace{-0.1in}
\caption{Examples of video editing (i.e., add object, replace object and style transfer) results by different approaches. }
\label{fig:comparison_1}
\vspace{-0.1in}
\end{figure*}

\begin{figure}
	\centering
	\includegraphics[width=0.98\linewidth]{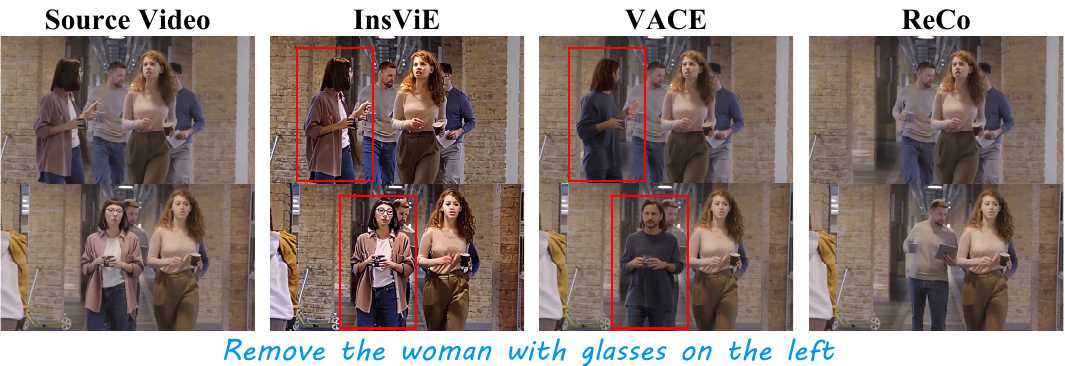}
    \vspace{-0.1in}
	\caption{Visual comparisons on the object removal task.}
    \vspace{-0.2in}
	\label{fig:comparison_2}
\end{figure}


\subsection{Comparisons with State-of-the-Art Methods}
We compare our ReCo with several state-of-the-art instructional video editing methods, including InsViE~\cite{insvie2025}, Ditto~\cite{ditto2025}, Lucy-Edit~\cite{decart2025lucyedit}, and VACE~\cite{jiang2025vace}, on our VLLM-based benchmark. 
Table~\ref{tab:tasks-methods-final} summarizes the performance comparisons on the four video editing tasks. 
Overall, ReCo consistently outperforms existing baselines on the total score $S$ across all tasks.
In particular, for the local video editing, ReCo attains the total score $S$ of $8.23$ on \textit{Add} and $8.74$ on \textit{Replace}, surpassing the strong competitor Ditto ($7.56$) and Lucy-Edit ($6.72$) by $0.67$ and $2.02$, respectively. 
Significant performance trends can also be observed on the editing accuracy perspective (i.e., $S_{EA}$).
The results demonstrate that our ReCo not only accurately follows the instruction prompt to correctly localize the editing region but also preserves the contents of non-edited areas.
In terms of video naturalness (i.e., $S_{VN}$), the better performances achieved by our model verifies the efficacy of naturally integrate editing objects into source video.
Although the $S_{VN}$ of ReCo is slightly below that of Ditto on the \textit{Add} task, ours can better keep original video contents while Ditto tends to re-render the whole video into different color style as shown in Figure \ref{fig:comparison_1}.
The phenomenon is also evidenced by the lower $S_{EA}$ score ($6.70$) of Ditto.
Additionally, the best performance of video quality ($S_{VQ}$) further indicates that the videos generated by ReCo have minimal visual artifacts or degradation.
Even under the multi-task training setting (i.e., unify local editing and global stylization) that could bring some conflicts during model optimization, ReCo still manifests the strong capability for video style transfer and attains $9.17$ of the total score $S$. 
All these results basically validate the merit of performing regional constraint modeling on in-context generation for instructional video editing. 



\begin{table*}[t]
    \centering
    \caption{Performance comparisons among different variants of ReCo on four video editing tasks.} 
    
    \label{tab:ablation_table}
    \vspace{-5pt}
    \setlength{\tabcolsep}{0.7em}
    \resizebox{\textwidth}{!}{%
    \begin{tabular}{l | cccc | cccc | cccc | cccc}
        \toprule
        \multirow{2}{*}{\textbf{Model}} &
        \multicolumn{4}{c|}{\textbf{Add}} &
        \multicolumn{4}{c|}{\textbf{Replace}} &
        \multicolumn{4}{c|}{\textbf{Remove}} &
        \multicolumn{4}{c}{\textbf{Style}} \\
        \cmidrule{2-17}
        & $S_{EA}$ & $S_{VN}$ & $S_{VQ}$ & $S$
        & $S_{EA}$ & $S_{VN}$ & $S_{VQ}$ & $S$
        & $S_{EA}$ & $S_{VN}$ & $S_{VQ}$ & $S$
        & $S_{EA}$ & $S_{VN}$ & $S_{VQ}$ & $S$ \\
        \midrule
        ReCo$_{LC-}$  & 8.05 & \underline{7.44} & \underline{8.59} & \underline{8.03} & 9.01 & \underline{8.01} & \underline{8.67} & \underline{8.56} & 6.90 & \underline{6.83} & \textbf{6.91} & \underline{6.88} & 9.09 & \underline{9.10} & \underline{8.84} & 9.01 \\
        ReCo$_{AC-}$  & \underline{8.33} & 7.37 & 8.01 & 7.90 & \underline{9.23} & 7.94 & 8.46 & 8.54 & \underline{7.11} & 6.75 & 6.70 & 6.85 & \underline{9.21} & 9.08 & 8.81 & \underline{9.03} \\
        ReCo   & \textbf{8.54} & \textbf{7.55} & \textbf{8.61} & \textbf{8.23} & \textbf{9.43} & \textbf{8.01} & \textbf{8.77} & \textbf{8.74} & \textbf{7.28} & \textbf{6.90} & \underline{6.82} & \textbf{7.00} & \textbf{9.43} & \textbf{9.19} & \textbf{8.90} & \textbf{9.17} \\
        \bottomrule
    \end{tabular}}
    \vspace{-0.18in}
\end{table*}

Figure~\ref{fig:comparison_1} and~\ref{fig:comparison_2} further show the video editing results on the four tasks. 
Generally, compared to other baselines, ReCo edits videos with better instruction following, higher video quality and better background consistency. 
For instance, InsViE tends to produce videos with artifacts and usually suffers from editing failure. 
Recent Lucy-Edit exhibits poor instruction-following and fails to accurately render the specified attributes (e.g., brown chimpanzee wearing a hoodie).
Though Ditto generates natural-looking objects in the \textit{Add} task, it struggles to preserve background consistency of non-editing regions and localize the accurate editing region (e.g., adding the crown at the back of the seal).
Meanwhile, the ability of instruction following for Ditto is inferior to ours, erroneously synthesizing a new monkey alongside the man instead of replacing him.
We speculate that these issues of Ditto are caused by the lack of regional correlation modeling for in-context generation when directly fine-tuning VACE~\cite{jiang2025vace} with textual instructions.
Our ReCo, in comparison, regulates the in-context generation learning with the region-wise constraints to emphasize editing region localization and alleviate cross-region token interference simultaneously.
Thus, the videos modified by ReCo reflect both accurate editing results and natural novel object integration with the original video background.



\begin{figure}
	\centering
	\includegraphics[width=1.0\linewidth]{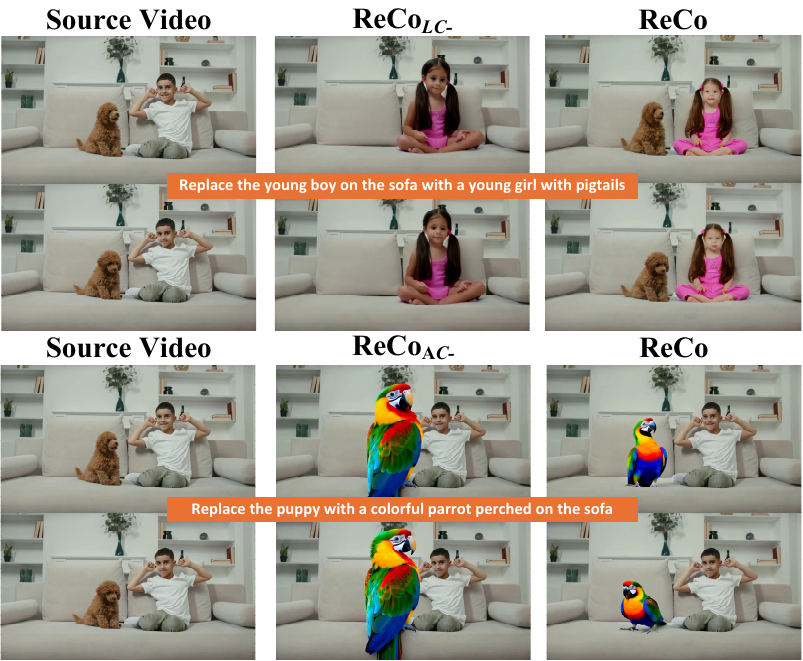}
	\vspace{-0.20in}
	\caption{Editing results on replace task among variants of ReCo.}
	\vspace{-0.20in}
	\label{fig:ablation}
\end{figure}

\subsection{Ablation Study on Regional Constraint}
We investigate how the two regional constraints in our ReCo influence the final instruction-based video editing. 
Table~\ref{tab:ablation_table} summarizes the video editing performances of different variants of our ReCo.
Two additional runs are involved, i.e., ReCo$_{LC-}$ and ReCo$_{AC-}$, which remove the latent and attention regional constraint in ReCo, respectively.
Specifically, when the region constraint in latent space is discarded, there is a dramatic performance drop on $S_{EA}$, which indicates a significant decay of editing accuracy.
The scores of $S_{VN}$ and $S_{VQ}$ also decrease slightly but remain comparable. 
The results highlight the effectiveness of latent region constraint learning to amplify accurate localization of editing region.
The top part of Figure~\ref{fig:ablation} further visualizes one video editing case among ReCo$_{LC-}$ and ReCo. 
Given the instruction of ``replace the young boy on the sofa with a young girl with pigtails,'' ReCo$_{LC-}$ could replace the boy but incorrectly removes the nearby dog. 

When removing the regularization term in attention space, ReCo$_{AC-}$ performs worse on the video naturalness perspective (i.e., $S_{VN}$) as shown in Table~\ref{tab:ablation_table}.
We also show one editing example in the lower part of Figure~\ref{fig:ablation}.
As shown in the figure, ReCo$_{AC-}$ generates a big parrot which has an unnatural scale relative to the environment. 
With the equipment of attention regularization that reduces the token interference from editing area and strengthens the interaction with background in novel object generation, ReCo synthesizes the parrot with natural size and better coherence.

\section{Conclusions}
We have presented ReCo that shapes in-context generation for instruction-based video editing.
Particularly, we study the problem of integrating the regional constraint modeling between editing and non-editing areas into diffusion training. 
To materialize our idea, ReCo jointly denoises the width-concatenated source-target video pair based on the natural language instructions, and conducts two regularization terms to emphasize region-wise relationship on both one-step backward denoised latents and attention maps.
To alleviate unexpected content generation in non-editing regions, the regularization term in latent space tries to decrease the latent discrepancy of non-editing regions between source and target videos, while increasing the differences at the editing area.
Meanwhile, ReCo suppresses the attention of tokens in the editing region to tokens in the same part of source video, which alleviates the interference from original editing region tokens to novel object generation.
Moreover, we carefully construct a high-quality video editing dataset, i.e., ReCo-Data, consisting of 500K instruction-video pairs covering a wide range of editing tasks.
Extensive experiments across four editing tasks verify the superiority of ReCo over state-of-the-art approaches.

{
    \small
    \bibliographystyle{ieeenat_fullname}
    \bibliography{main}

@String(CVPR= {IEEE Conf. Comput. Vis. Pattern Recog.})

@String(ICCV= {Int. Conf. Comput. Vis.})

@String(ECCV= {Eur. Conf. Comput. Vis.})

@String(ICLR = {Int. Conf. Learn. Represent.})

@String(AAAI = {AAAI})

@String(CVPR  = {CVPR})

@String(ICCV  = {ICCV})

@String(ECCV  = {ECCV})

@String(ICLR  = {ICLR})

@article{wan2025,
      title={{Wan: Open and Advanced Large-Scale Video Generative Models}}, 
      author={Team Wan and Ang Wang and Baole Ai and Bin Wen and Chaojie Mao and Chen-Wei Xie and Di Chen and Feiwu Yu and Haiming Zhao and Jianxiao Yang and others},
      journal = {arXiv preprint arXiv:2503.20314},
      year={2025}
}

@article{iclora2024,
	title={{In-Context LoRA for Diffusion Transformers}},
	author={Lianghua Huang and Wei Wang and Zhi-Fan Wu and Yupeng Shi and Huanzhang Dou and Chen Liang and Yutong Feng, Yu Liu and Jingren Zhou},
	journal={arXiv preprint arxiv:2410.23775},
	year={2024}
}

@article{kontext2025,
	title={{FLUX.1 Kontext: Flow Matching for In-Context Image Generation and Editing in Latent Space}},
	author={Black Forest Labs and Stephen Batifol and Andreas Blattmann and Frederic Boesel and Saksham Consul and Cyril Diagne and Tim Dockhorn and Jack English and Zion English and Patrick Esser and Sumith Kulal and others},
	journal={arXiv:2506.15742},
	year={2025}
}

@article{team2023gemini,
  title={{Gemini: A Family of Highly Capable Multimodal Models}},
  author={Gemini Team Google: Rohan Anil and Sebastian Borgeaud and Jean-Baptiste Alayrac and Jiahui Yu and Radu Soricut and Johan Schalkwyk and Andrew M. Dai, Anja Hauth and Katie Millican and others},
  journal={arXiv preprint arXiv:2312.11805},
  year={2023}
}

@article{HunyuanCustom2025,
	title={{HunyuanCustom: A Multimodal-Driven Architecture for Customized Video Generation}},
	author={Teng Hu and Zhentao Yu and Zhengguang Zhou and Sen Liang and Yuan Zhou and Qin Lin and Qinglin Lu},
	journal={arXiv preprint arXiv:2505.04512},
	year={2025}
}

@article{vae,
	title={{Auto-Encoding Variational Bayes}},
	author={Diederik P Kingma and Max Welling},
	journal={arXiv preprint arXiv:1312.6114},
	year={2013}
}

@article{ditto2025,
	title={{Scaling Instruction-Based Video Editing with a High-Quality Synthetic Dataset}},
	author={Qingyan Bai and Qiuyu Wang and Hao Ouyang and Yue Yu and Hanlin Wang and Wen Wang and Ka Leong Cheng, Shuailei Ma and Yanhong Zeng and Zichen Liu and Yinghao Xu and Yujun Shen and Qifeng Chen},
	journal={arXiv:2510.15742},
	year={2025}
}

@article{qwenimage2025,
	title={{Qwen-Image Technical Report}},
	author={Chenfei Wu and Jiahao Li and Jingren Zhou, Junyang Lin and Kaiyuan Gao and Kun Yan and Sheng-ming Yin and Shuai Bai and Xiao Xu and Yilei Chen and Yuxiang Chen and others},
	journal={arXiv:2508.02324},
	year={2025}
}

@article{qwenvl2025,
	title={{Qwen3-VL Technical Report}},
	author={Shuai Bai and Yuxuan Cai and Ruizhe Chen and Keqin Chen and Xionghui Chen and Zesen Cheng and Lianghao Deng and Wei Ding and Chang Gao and Chunjiang Ge and others},
	journal={arXiv:2511.21631},
	year={2025}
}

@article{decart2025lucyedit,
  title   = {{Lucy Edit: Open-Weight Text-Guided Video Editing}},
  author  = {DecartAI Team},
  year    = {2025},
 }

@article{flowdirector2025,
	title={{FlowDirector: Training-Free Flow Steering for Precise Text-to-Video Editing}},
	author={Guangzhao Li and Yanming Yang and Chenxi Song and Chi Zhang},
	journal={arXiv:2506.05046},
	year={2025}
}

@article{2024lumiere,
	title={{Lumiere: A Space-Time Diffusion Model for Video Generation}},
	author={Omer Bar-Tal and Hila Chefer and Omer Tov and Charles Herrmann and Roni Paiss and Shiran Zada and Ariel Ephrat and Junhwa Hur and Guanghui Liu and Amit Raj and Yuanzhen Li and Michael Rubinstein and Tomer Michaeli and Oliver Wang and Deqing Sun and Tali Dekel and Inbar Mosseri},
	journal={arXiv preprint arXiv:2401.12945},
	year={2024}
}

@article{sora2024,
	title={{Video Generation Models as World Simulators}},
	author={Tim Brooks and Bill Peebles and Connor Holmes and Will DePue and Yufei Guo and Li Jing and David Schnurr and Joe Taylor and Troy Luhman and Eric Luhman and Clarence Ng and Ricky Wang and Aditya Ramesh},
	year={2024},
}

@article{2024vidu,
	title={{Vidu: a Highly Consistent, Dynamic and Skilled Text-to-Video Generator with Diffusion Models}},
	author={Fan Bao and Chendong Xiang and Gang Yue and Guande He and Hongzhou Zhu and Kaiwen Zheng and Min Zhao and Shilong Liu and Yaole Wang and Jun Zhu},
	journal={arXiv preprint arXiv:2405.04233},
	year={2024}
}

@article{2024hunyuanvideo,
	title={{HunyuanVideo: A Systematic Framework For Large Video Generative Models}},
	author={Weijie Kong and Qi Tian and Zijian Zhang and Rox Min and Zuozhuo Dai and Jin Zhou and Jiangfeng Xiong and Xin Li and Bo Wu and Jianwei Zhang and Kathrina Wu and Qin Lin and Junkun Yuan and Yanxin Long and Aladdin Wang and Andong Wang and others},
	journal={arXiv preprint arXiv:2412.03603},
	year={2024}
}

@article{HaCohen2024LTXVideo,
  title={{LTX-Video: Realtime Video Latent Diffusion}},
  author={HaCohen, Yoav and Chiprut, Nisan and Brazowski, Benny and Shalem, Daniel and Moshe, Dudu and Richardson, Eitan and Levin, Eran and Shiran, Guy and Zabari, Nir and Gordon, Ori and Panet, Poriya and Weissbuch, Sapir and Kulikov, Victor and Bitterman, Yaki and Melumian, Zeev and Bibi, Ofir},
  journal={arXiv preprint arXiv:2501.00103},
  year={2025}
}

@article{2025waver,
	title={{Waver: Wave Your Way to Lifelike Video Generation}},
	author={Yifu Zhang and Hao Yang and Yuqi Zhang and Yifei Hu and Fengda Zhu and Chuang Lin and Xiaofeng Mei and Yi Jiang and Bingyue Peng and Zehuan Yuan},
	journal={arXiv preprint arXiv:2508.15761},
	year={2025}
}

@article{midas,
    title     = {{Towards Robust Monocular Depth Estimation: Mixing Datasets for Zero-shot Cross-dataset Transfer}},
    author    = {René Ranftl and Katrin Lasinger and David Hafner and Konrad Schindler and Vladlen Koltun},
    journal = {IEEE TPAMI},
    year      = {2020}
    }

@article{zhao2025ObjectClear,
    title     = {{ObjectClear: Complete Object Removal via Object-Effect Attention}},
    author    = {Zhao Jixin and Zhou Shangchen and Wang Zhouxia and Yang Peiqing and Loy Chen Change},
    journal = {arXiv preprint arXiv:2505.22636},
    year      = {2025}
    }

@article{bai2025qwen2,
  title={Qwen2. 5-vl technical report},
  author={Shuai Bai and Keqin Chen and Xuejing Liu and Jialin Wang and Wenbin Ge and Sibo Song and Kai Dang and Peng Wang and Shijie Wang and Jun Tang and others},
  journal={arXiv preprint arXiv:2502.13923},
  year={2025}
}

@article{cui2025paddleocr30technicalreport,
      title={{PaddleOCR 3.0 Technical Report}}, 
      author={Cheng Cui and Ting Sun and Manhui Lin and Tingquan Gao and Yubo Zhang and Jiaxuan Liu and Xueqing Wang and Zelun Zhang and Changda Zhou and Hongen Liu and Yue Zhang and Wenyu Lv and Kui Huang and Yichao Zhang and Jing Zhang and Jun Zhang and Yi Liu and Dianhai Yu and Yanjun Ma},
      journal={arXiv preprint arXiv:2507.05595},
      year={2025},
}

@article{pyscenedetect,
  title={{PySceneDetect}},
  author={PySceneDetect Developers},
  year={2024},
  url={https://www.scenedetect.com},
}

@article{spaCy,
  title={{spaCy}},
  author={spaCy Developers},
  year={2024},
  url={https://github.com/explosion/spaCy},
}

@article{improved-aesthetic-predictor,
  title={{improved-aesthetic-predictor}},
  author={christophschuhmann},
  year={2024},
  journal={improved-aesthetic-predictor Lab},
  url={https://github.com/christophschuhmann/improved-aesthetic-predictor/tree/main},
}

@article{groundsam,
  title={{Grounded SAM: Assembling Open-World Models for Diverse Visual Tasks}},
  author={Tianhe Ren and Shilong Liu and Ailing Zeng and Jing Lin and Kunchang Li and He Cao and Jiayu Chen and Xinyu Huang and Yukang Chen and Feng Yan and Zhaoyang Zeng and Hao Zhang and Feng Li and Jie Yang and Hongyang Li and Qing Jiang and Lei Zhang},
  journal={arXiv preprint arXiv:2401.14159},
  year={2024}
}

@article{videofactory,
  title={{VideoFactory: Swap Attention in Spatiotemporal Diffusions for Text-to-Video Generation}},
  author={Wenjing Wang and Huan Yang and Zixi Tuo and Huiguo He and Junchen Zhu and Jianlong Fu and Jiaying Liu},
  journal={arXiv preprint arXiv:2305.10874},
  year={2023}
}

@article{liu2025javisdit,
  title={{JavisDiT: Joint Audio-Video Diffusion Transformer with Hierarchical Spatio-Temporal Prior Synchronization}},
  author={Kai Liu and Wei Li and Lai Chen and Shengqiong Wu and Yanhao Zheng and Jiayi Ji and Fan Zhou and Rongxin Jiang and Jiebo Luo and Hao Fei and Tat-Seng Chua},
  journal={arXiv preprint arXiv:2503.23377},
  year={2025}
}

@article{yuan2025opens2v,
  title={{OpenS2V-Nexus: A Detailed Benchmark and Million-Scale Dataset for Subject-to-Video Generation}},
  author={Shenghai Yuan and Xianyi He and Yufan Deng and Yang Ye and Jinfa Huang and Bin Lin and Jiebo Luo and Li Yuan},
  journal={arXiv preprint arXiv:2505.20292},
  year={2025}
}

@inproceedings{zichengedit,
    title = {{Towards Consistent Video Editing with Text-to-Image Diffusion Models}},
    author = {Zicheng Zhang and Bonan Li and Xuecheng Nie and Congying Han and Tiande Guo and Luoqi Liu},
    booktitle = {NeurIPS},
    year = {2023}
}

@inproceedings{followpose,
    title = {{Follow Your Pose: Pose-Guided Text-to-Video Generation using Pose-Free Videos}},
    author = {Yue Ma and Yingqing He and Xiaodong Cun and Xintao Wang and Siran Chen and Ying Shan and Xiu Li and Qifeng Chen},
    booktitle = {AAAI},
    year = {2023}
}

@inproceedings{raft,
    title = {{RAFT: Recurrent All-Pairs Field Transforms for Optical Flow}},
    author = {Zachary Teed and Jia Deng},
    booktitle = {ECCV},
    year = {2020}
}

@inproceedings{2025motionpro,
 title={{MotionPro: A Precise Motion Controller for Image-to-Video Generation}},
 author={Zhongwei Zhang and Fuchen Long and Zhaofan Qiu and Yingwei Pan and Wu Liu and Ting Yao and Tao Mei},
 booktitle={CVPR},
 year={2025}
}

@inproceedings{zhang2024trip,
  title={{TRIP: Temporal Residual Learning with Image Noise Prior for Image-to-Video Diffusion Models}},
  author={Zhongwei Zhang and Fuchen Long and Yingwei Pan and Zhaofan Qiu and Ting Yao and Yang Cao and Tao Mei},
  booktitle={CVPR},
  year={2024}
}

@inproceedings{sddit,
 title={{SD-DiT: Unleashing the Power of Self-supervised Discrimination in Diffusion Transformer}},
 author={Rui Zhu and Yingwei Pan and Yehao Li and Ting Yao and Zhenglong Sun and Tao Mei and Chang Wen Chen},
 booktitle={CVPR},
 year={2025}
}

@inproceedings{sam2,
    title = {{SAM 2: Segment Anything in Images and Videos}},
    author = {Ravi Nikhila and Gabeur Valentin and Hu Yuan-Ting and Hu Ronghang and Ryali Chaitanya and Ma Tengyu and Khedr Haitham and R{"a}dle Roman and Rolland Chloe and Gustafson Laura and Mintun Eric and Pan Junting and Alwala Kalyan Vasudev and Carion Nicolas and Wu Chao-Yuan and Girshick Ross and Doll{'a}r Piotr and Feichtenhofer Christoph},
    booktitle = {ICLR},
    year = {2025}
}

@inproceedings{2024cogvideox,
	title={{CogVideoX: Text-to-Video Diffusion Models with An Expert Transformer}},
	author={Zhuoyi Yang and Jiayan Teng and Wendi Zheng and Ming Ding and Shiyu Huang and Jiazheng Xu and Yuanming Yang and Wenyi Hong and Xiaohan Zhang and Guanyu Feng and Da Yin and Xiaotao Gu and Yuxuan Zhang and Weihan Wang and Yean Cheng and Ting Liu and Bin Xu and Yuxiao Dong and Jie Tang},
  	booktitle={ICLR},
	year={2025}
}

@inproceedings{genprop2025,
	title={{Generative Video Propagation}},
	author={Shaoteng Liu and Tianyu Wang and Jui-Hsien Wang and Qing Liu and Zhifei Zhang and Joon-Young Lee, Yijun Li and Bei Yu and Zhe Lin and Soo Ye Kim and Jiaya Jia},
	booktitle={CVPR},
	year={2025},
}

@inproceedings{flatten2024,
	title={{FLATTEN: optical FLow-guided ATTENtion for consistent text-to-video editing}},
	author={Yuren Cong and Mengmeng Xu and Christian Simon and Shoufa Chen and Jiawei Ren and Yanping Xie and Juan-Manuel Perez-Rua and Bodo Rosenhahn and Tao Xiang and Sen He},
	booktitle={ICLR},
	year={2024},
}

@inproceedings{videotome2024,
	title={{VidToMe: Video Token Merging for Zero-Shot Video Editing}},
	author={Xirui Li and Chao Ma and Xiaokang Yang and Ming-Hsuan Yang},
	booktitle={CVPR},
	year={2024},
}

@inproceedings{videop2p2024,
	title={{Video-P2P: Video Editing with Cross-attention Control}},
	author={Shaoteng Liu and Yuechen Zhang and Wenbo Li and Zhe Lin and Jiaya Jia},
	booktitle={CVPR},
	year={2024},
}

@inproceedings{2023tune,
	title={{Tune-A-Video: One-Shot Tuning of Image Diffusion Models for Text-to-Video Generation}},
	author={Jay Zhangjie Wu and Yixiao Ge and Xintao Wang and Weixian Lei and Yuchao Gu and Yufei Shi and Wynne Hsu and Ying Shan and Xiaohu Qie and Mike Zheng Shou},
	booktitle={ICCV},
	year={2023},
}

@inproceedings{FlowEdit,
	title={{FlowEdit: Inversion-Free Text-Based Editing Using Pre-Trained Flow Models}},
	author={Vladimir Kulikov and Matan Kleiner and Inbar Huberman-Spiegelglas and Tomer Michaeli},
	booktitle={ICCV},
	year={2025}
}

@inproceedings{anyv2v2014,
	title={{AnyV2V: A Tuning-Free Framework For Any Video-to-Video Editing Tasks}},
	author={Max Ku and Cong Wei and Weiming Ren and Harry Yang and Wenhu Chen},
	booktitle={TMLR},
	year={2024}
}

@inproceedings{tokenflow2024,
	title={{TokenFlow: Consistent Diffusion Features for Consistent Video Editing}},
	author={Michal Geyer and Omer Bar-Tal and Shai Bagon and Tali Dekel},
	booktitle={ICLR},
	year={2024}
}

@inproceedings{rave2024,
	title={{RAVE: Randomized Noise Shuffling for Fast and Consistent Video Editing with Diffusion Models}},
	author={Ozgur Kara and Bariscan Kurtkaya and Hidir Yesiltepe and James M. Rehg and Pinar Yanardag},
	booktitle={CVPR},
	year={2024}
}

@inproceedings{2024SDXL,
	title={{SDXL: Improving Latent Diffusion Models for High-Resolution Image Synthesis}},
	author={Dustin Podell and Zion English and Kyle Lacey and Andreas Blattmann and Tim Dockhorn and Jonas Müller and Joe Penna and Robin Rombach},
	booktitle={ICLR},
	year={2024}
}

@inproceedings{2023adding,
	title={{Adding Conditional Control to Text-to-Image Diffusion Models}},
	author={Lvmin Zhang and Anyi Rao and Maneesh Agrawala},
	booktitle={ICCV},
	year={2023}
}

@inproceedings{2023fatezero,
	title={{FateZero: Fusing Attentions for Zero-Shot Text-Based Video Editing}},
	author={Chenyang Qi and Xiaodong Cun and Yong Zhang and Chenyang Lei and Xintao Wang and Ying Shan and Qifeng Chen},
	booktitle={ICCV},
	year={2023}
}

@inproceedings{2022prompt,
	title={{Prompt-to-Prompt Image Editing with Cross-Attention Control}},
	author={Amir Hertz and Ron Mokady and Jay Tenenbaum and Kfir Aberman and Yael Pritch and Daniel Cohen-or},
	booktitle={ICLR},
	year={2023}
}

@inproceedings{2020ddpm,
	title={{Denoising Diffusion Probabilistic Models}},
	author={Jonathan Ho and Ajay Jain and Pieter Abbeel},
	booktitle={NeurIPS},
	year={2020}
}

@inproceedings{2023instructpix2pix,
	title={{InstructPix2Pix: Learning to Follow Image Editing Instructions}},
	author={Tim Brooks and Aleksander Holynski and Alexei A. Efros},
	booktitle={CVPR},
	year={2023}
}

@inproceedings{2024emuedit,
	title={{Emu Edit: Precise Image Editing via Recognition and Generation Tasks}},
	author={Shelly Sheynin and Adam Polyak and Uriel Singer and Yuval Kirstain and Amit Zohar and Oron Ashual and Devi Parikh and Yaniv Taigman},
	booktitle={CVPR},
	year={2024}
}

@article{2024hidreami1,
	title={{HiDream-I1: A High-Efficient Image Generative Foundation Model with Sparse Diffusion Transformer}},
	author={Qi Cai and Jingwen Chen and Yang Chen and Yehao Li and Fuchen Long and Yingwei Pan and Zhaofan Qiu and others},
	journal={arXiv preprint arXiv:2505.22705},
	year={2025}
}

@inproceedings{long2024eccv,
  title={{VideoStudio: Generating Consistent-Content and Multi-Scene Videos}},
  author={Fuchen Long and Zhaofan Qiu and Ting Yao and Tao Mei},
  booktitle={ECCV},
  year={2024}
}

@inproceedings{2022ldm,
	title={{High-Resolution Image Synthesis with Latent Diffusion Models}},
	author={Robin Rombach and Andreas Blattmann and Dominik Lorenz and Patrick Esser and Björn Ommer},
	booktitle={CVPR},
	year={2022}
}

@inproceedings{2021glide,
	title={{GLIDE: Towards Photorealistic Image Generation and Editing with Text-Guided Diffusion Models}},
	author={Alex Nichol and Prafulla Dhariwal and Aditya Ramesh and Pranav Shyam and Pamela Mishkin and Bob McGrew and Ilya Sutskever and Mark Chen},
	booktitle={ICML},
	year={2022}
}

@inproceedings{ominiedit2025,
  title={{OmniEdit: Building Image Editing Generalist Models Through Specialist Supervision}},
  author={Cong Wei and Zheyang Xiong and Weiming Ren, Xinrun Du and Ge Zhang and Wenhu Chen},
  booktitle={ICLR},
  year={2025}
}

@inproceedings{2022photorealistic,
	title={{Photorealistic Text-to-Image Diffusion Models with Deep Language Understanding}},
	author={Chitwan Saharia and William Chan and Saurabh Saxena and Lala Li and Jay Whang and Emily Denton and Seyed Kamyar Seyed Ghasemipour and Burcu Karagol Ayan and S. Sara Mahdavi and Rapha Gontijo Lopes and Tim Salimans and Jonathan Ho and David J Fleet and Mohammad Norouzi},
	booktitle={NeurIPS},
	year={2022}
}

@inproceedings{senorita2025,
  title={{Señorita-2M: A High-Quality Instruction-based Dataset for General Video Editing by Video Specialists}},
  author={Bojia Zi and Penghui Ruan and Marco Chen, Xianbiao Qi and Shaozhe Hao and Shihao Zhao and Youze Huang and Bin Liang and Rong Xiao and Kam-Fai Wong},
  booktitle={NeurIPS},
  year={2025}
}

@inproceedings{insv2v2024,
  title={{Consistent Video-to-Video Transfer Using Synthetic Dataset}},
  author={Jiaxin Cheng and Tianjun Xiao and Tong He},
  booktitle={ICLR},
  year={2024}
}

@inproceedings{insvie2025,
  title={{InsViE-1M: Effective Instruction-based Video Editing with Elaborate Dataset Construction}},
  author={Yuhui Wu and Liyi Chen and Ruibin Li and Shihao Wang and Chenxi Xie and Lei Zhang},
  booktitle={ICCV},
  year={2025}
}

@inproceedings{videopainter2025,
	title={{VideoPainter: Any-length Video Inpainting and Editing with Plug-and-Play Context Control}},
	author={Yuxuan Bian and Zhaoyang Zhang and Xuan Ju and Mingdeng Cao and Liangbin Xie and Ying Shan and Qiang Xu},
	booktitle={CVPR},
	year={2025}
}

@inproceedings{flowmatch2022,
  title={{Flow Matching for Generative Modeling}},
  author={Yaron Lipman and Ricky T. Q. Chen and Heli Ben-Hamu and Maximilian Nickel and Matt Le},
  booktitle={ICLR},
  year={2023}
}

@inproceedings{ominigen2025,
  title={{OmniGen: Unified Image Generation}},
  author={Shitao Xiao and Yueze Wang and Junjie Zhou and Huaying Yuan and Xingrun Xing and Ruiran Yan and Chaofan Li and Shuting Wang and Tiejun Huang and Zheng Liu},
  booktitle={CVPR},
  year={2025}
}

@inproceedings{flowmatchldm2024,
  title={{Scaling Rectified Flow Transformers for High-Resolution Image Synthesis}},
  author={Patrick Esser and Sumith Kulal and Andreas Blattmann and Rahim Entezari and Jonas Müller and Harry Saini and Yam Levi and Dominik Lorenz and Axel Sauer and Frederic Boesel and others},
  booktitle={ICML},
  year={2024}
}

@inproceedings{icedit2025,
	title={{Enabling Instructional Image Editing with In-Context Generation in Large Scale Diffusion Transformer}},
	author={Zechuan Zhang and Ji Xie and Yu Lu and Zongxin Yang and Yi Yang},
	booktitle={NeurIPS},
	year={2025}
}

@inproceedings{ominicontrol2025,
	title={{OminiControl: Minimal and Universal Control for Diffusion Transformer}},
	author={Zhenxiong Tan and Songhua Liu and Xingyi Yang and Qiaochu Xue and Xinchao Wang},
	booktitle={ICCV},
	year={2025}
}

@inproceedings{Diptych2025,
	title={{Large-Scale Text-to-Image Model with Inpainting is a Zero-Shot Subject-Driven Image Generator}},
	author={Chaehun Shin and Jooyoung Choi and Heeseung Kim and Sungroh Yoon},
	booktitle={CVPR},
	year={2025}
}

@inproceedings{zhang2025icedit,
	title={{In-Context Edit: Enabling Instructional Image Editing with In-Context Generation in Large Scale Diffusion Transformer}},
	author={Zechuan Zhang and Ji Xie and Yu Lu and Zongxin Yang and Yi Yang},
	booktitle={NeurIPS},
	year={2025}
}

@inproceedings{jiang2025vace,
    title = {{VACE: All-in-One Video Creation and Editing}},
    author = {Jiang, Zeyinzi and Han, Zhen and Mao, Chaojie and Zhang, Jingfeng and Pan, Yulin and Liu, Yu},
    booktitle = {ICCV},
    year = {2025}
}

@inproceedings{hu2021region,
	title={{Region-Aware Contrastive Learning for Semantic Segmentation}},
	author={Hu, Hanzhe and Cui, Jinshi and Wang, Liwei},
	booktitle={ICCV},
	year={2021}
}

@inproceedings{2022imagenvideo,
	title={{Imagen Video: High Definition Video Generation with Diffusion Models}},
	author={Jonathan Ho and William Chan and Chitwan Saharia and Jay Whang and Ruiqi Gao and Alexey Gritsenko and Diederik P. Kingma and Ben Poole and Mohammad Norouzi and David J. Fleet and Tim Salimans},
	booktitle={CVPR},
	year={2022}
}

@inproceedings{2023align,
	title={{Align your Latents: High-Resolution Video Synthesis with Latent Diffusion Models}},
	author={Andreas Blattmann and Robin Rombach and Huan Ling and Tim Dockhorn and Seung Wook Kim and Sanja Fidler and Karsten Kreis},
	booktitle={CVPR},
	year={2023}
}

@inproceedings{2023make-a-video,
	title={{Make-a-Video: Text-to-Video Generation without Text-Video Data}},
	author={Uriel Singer and Adam Polyak and Thomas Hayes and Xi Yin and Jie An and Songyang Zhang and Qiyuan Hu and Harry Yang and Oron Ashual and Oran Gafni and Devi Parikh and Sonal Gupta and Yaniv Taigman},
	booktitle={ICLR},
	year={2023}
}

@inproceedings{2021ddimp,
	title={{Denoising Diffusion Implicit Models}},
	author={Jiaming Song and Chenlin Meng and Stefano Ermon},
	booktitle={ICLR},
	year={2021},
}
}

\newpage
\clearpage
\setcounter{page}{1}
\setcounter{section}{0}

\twocolumn[
{
\begin{center}
    {\Large \textbf{Region-Constraint In-Context Generation for Instructional Video Editing}}\\[1em]
    {\large \textbf{--- Supplementary Material}}
\end{center}
\vspace{1em}
}
]

The supplementary material contains: 1) the construction pipeline of ReCo-Data; 2) the details of VLLM-based benchmark for evaluation; 3) the implementation details of baselines and ReCo; 4) the generalization ability of ReCo.

\section{Construction Pipeline of ReCo-Data} \label{sec: reco-data}
Though instructional video editing has seen remarkable advances recently, the absence of large-scale, high-quality training datasets remains a critical hurdle. To overcome this, we present \textbf{ReCo-Data}, a dataset carefully designed to facilitate four key editing tasks: instance-level object addition, removal, replacement, and global video stylization. 

\begin{figure*}
\centering
\includegraphics[width=1.0\linewidth]{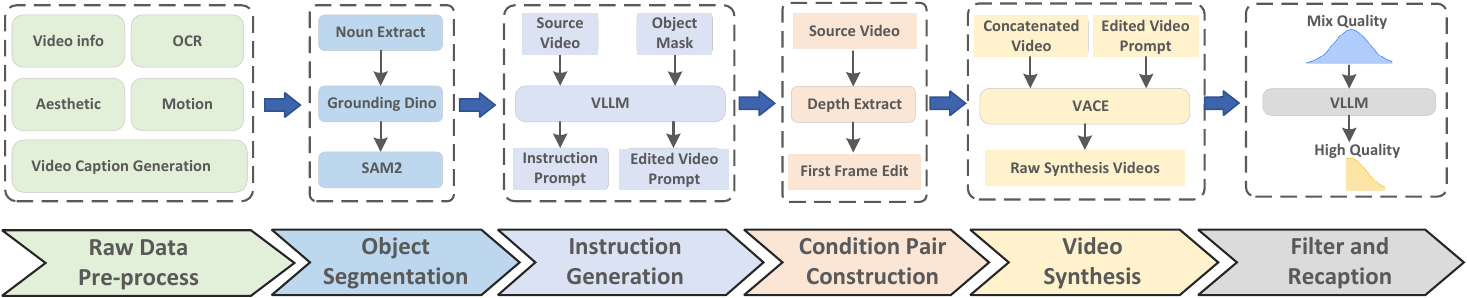}
\vspace{-0.2in}
\caption{An overview of our data construction pipeline. The process consists of six main stages: raw data pre-processing, object segmentation, instruction generation, condition pair construction, video synthesis, and video filtering and re-captioning.}

\label{fig:data_pipeline}
\vspace{-0.1in}
\end{figure*}

As illustrated in Figure~\ref{fig:data_pipeline}, the construction pipeline of ReCo-Data consists of six primary stages: (1) raw data pre-processing, where we filter raw video data based on specific quality criteria; (2) object segmentation, extracting object mask from videos; (3) instruction generation, employing VLLM (i.e., Gemini-2.5-Flash-Thinking~\cite{team2023gemini}) to construct editing prompts; (4) condition pair construction, which involves first frame editing and depth map generation to prepare the input conditions for VACE~\cite{jiang2025vace}; (5) video synthesis, employing VACE to generate videos based on conditions; and (6) video filtering and re-captioning, where VLLM (i.e., Gemini-2.5-Flash-Thinking~\cite{team2023gemini}) is leveraged again to filter out low-quality samples and re-caption remained videos.

\subsection{Raw Data Pre-processing}

\textbf{Data Collection.} 
To ensure data diversity, we collect raw videos from multiple sources, including the HD-VG~\cite{videofactory}, OpenS2V-Nexus~\cite{yuan2025opens2v}, and videos from the Pixel website~\cite{videopainter2025}. We employ PySceneDetect~\cite{pyscenedetect} to segment the long, multi-scene videos into shorter, manageable clips.

\textbf{Data Filtering.} 
We first filter clips based on basic metadata, retaining those with a duration exceeding $5$ seconds, a frame rate greater than $24$ fps, and a resolution of at least 720P. 
Then, we utilize aesthetic scores~\cite{improved-aesthetic-predictor} and optical flow~\cite{raft} to select videos characterized by high aesthetic quality and appropriate motion magnitude. 
Finally, to ensure visual purity, we employ PaddleOCR~\cite{cui2025paddleocr30technicalreport} for watermark detection, spatially cropping the frames to exclude any text detected with a confidence score exceeding 0.7.

\textbf{Video Captioning.} For subsequent object segmentation and editing prompt construction, we utilize Qwen2.5-VL-32B~\cite{bai2025qwen2} to obtain detailed descriptions of remained videos.

\subsection{Object Segmentation}
To enable precise instance-level object editing (e.g., replacement and removal) by using video inpainting models like VACE~\cite{jiang2025vace}, we need to first isolate the target objects. 
Given the complexity of scenes containing multiple objects, we adopt a systematic segmentation approach. 
First, we define a taxonomy and employ a Named Entity Recognition (NER) model, i.e., SpaCy~\cite{spacy}, to extract relevant entity nouns from video captions. 
Subsequently, we utilize Grounding Dino~\cite{groundsam} to detect objects and obtain their bounding boxes. To ensure the quality of the proposals, we apply Non-Maximum Suppression (NMS) to filter out duplicate boxes with overlaps exceeding $25\%$ and discard boxes that are disproportionately large or small. Finally, using the bounding boxes as prompts, we employ SAM 2~\cite{sam2} to generate mask sequences for the target objects.

\subsection{Instruction Generation}\label{sec:instgen}
The protocols to prepare editing prompts exhibit variations across local editing and global video stylization tasks.

\textbf{Local Editing.} For the local video editing tasks, we provide Gemini with a tuple containing original video caption and one representative key frame. In this key frame, the target object region is explicitly highlighted with a red convex hull. Guided by a finely tuned system prompt, Gemini is required to generate an appropriate editing instruction along with a target video caption describing the post-edit state.

\textbf{Video Stylization.} The process is analogous to local editing. We leverage Gemini's creative capabilities to brainstorm diverse stylization instructions and generate the corresponding target video descriptions.

\subsection{Condition Pair Construction}
In this stage, we leverage the full capabilities of existing models to construct optimal condition pairs, which are fed into VACE~\cite{jiang2025vace} for edited video generation.
For each editing task, specific strategy is used for condition generation.


\textbf{Object Removal.} 
Inputting the masked video (derived from object masks) and target prompt into VACE for edited video generation often fails to eliminate the object cleanly or leads to the hallucination of unexpected contents. 
To mitigate this issue, we adopt a two-stage approach for edited video generation. 
First, we employ ObjectClear~\cite{zhao2025ObjectClear} on the first frame to perform clean object removal. 
Next, we concatenate this edited frame with the subsequent masked video frames, which is then fed into VACE to perform video inpainting, yielding stable and high-quality object removal.

\textbf{Object Addition.} 
We treat this task as the inverse of object removal. Once a valid removal pair is generated, we simply swap the source and edited videos to create a corresponding training pair for the object addition task.

\textbf{Object Replacement.} 
VACE demonstrates robust performance on object replacement. Therefore, we simply feed the masked video sequence and the target video prompt into VACE to generate high-quality replacement results.

\textbf{Video Stylization.} 
Although VACE supports video stylization conditioned on depth maps, maintaining the content structure of the original video is not satisfactory. 
Thus, we employ a strategy similar to object removal. We first utilize FLUX.1 Kontext~\cite{kontext2025} to apply the style transfer on the first frame. 
Subsequently, we concatenate the edited frame with the depth map sequence (extracted via MiDaS~\cite{midas}) to serve as the input condition pair for VACE, thereby generating a temporally consistent stylized video. 

Specifically designed for video stylization, our pipeline addresses common artifacts in VACE-generated data, such as frame collapse, abrupt transitions, temporal inconsistency, and content distortion (e.g., facial deformations). To ensure high-quality output, we implement a two-stage strategy. First, we use Qwen-3-VL-Instant~\cite{qwenvl2025} to filter for smooth and stable videos, removing low-quality frames with severe artifacts or flickering. Second, we refine the selected videos using the 14B Wan-2.2-T2V~\cite{wan2025} model. These combined strategies enable the synthesis of stylized videos with significantly improved visual and temporal quality.

\subsection{Video Synthesis}
Once the requisite condition pairs are prepared, we execute VACE in large-scale batches to synthesize high-quality editing videos.
To maximize the utility of the synthesized data and ensure efficient construction, we design a data augmentation strategy to generate additional training pairs without extra computational cost.

\textbf{Reversible Replacement.} 
For the object replacement based on one source video, we treat such process as reversible.
By swapping the source and target videos, we effectively double the volume of the replacement data.

\textbf{Cross-Task Augmentation.}
In fact, the edited videos generated from object removal and replacement share the same clean background.
Therefore, the synthesized video from the replacement (containing a novel object) can be paired with the background video from the removal task. 
This allows us to construct new ``removal'' pairs (new object $\rightarrow$ background) and ``adding'' pairs (background $\rightarrow$ new object), effectively doubling the dataset size for both tasks.

Finally, we totally construct approximately 800K video pairs for the four editing tasks. 
The entire data synthesis process required approximately \textbf{76,800 GPU hours} on NVIDIA RTX 4090.

\subsection{Video Filtering and Re-captioning}
To pursue high quality of instruction-video pairs, we employ the VLLM, i.e., Gemini-2.5-Flash-Thinking, to evaluate and filter out low-quality samples in total 800K video pairs. 
We extract representative key frames from the source and edited videos, and concatenate them into a side-by-side layout to facilitate VLLM assessment.
The remained video pairs are re-captioned by VLLM. 
The entire caption process (including Sec.~\ref{sec:instgen}) incurred a total cost of approximately $\$13,600$. Finally, we construct ReCo-Data with 500K high-quality instruction-video pairs. 
Each video clip contains $81$ frames with the resolution of $480\times 832$ and duration of $5$ seconds.


\section{VLLM-based Evaluation Benchmark}
\label{sec:reco_benchmark}

Traditional video generation metrics often struggle to accurately assess the fidelity and quality of video editing. 
Inspired by recent image editing evaluation protocols~\cite{ominiedit2025}, we propose a VLLM-based evaluation benchmark to comprehensively and effectively assess video editing quality.


\textbf{Testing Data.}
We collect $480$ video-instruction pairs as the testing data, distributed evenly with 120 pairs for each of the four tasks
(i.e., object add, remove, replace, and video stylization). 
All source videos are collected from Pexels video platform. 
For local editing tasks (i.e., object add, remove and replace), we utilize Gemini-2.5-Flash-Thinking~\cite{team2023gemini} to brainstorm and generate diverse editing instructions based on the video content. 
For rigorous evaluation on video stylization, we randomly select $10$ source videos and apply $12$ distinct styles to each, resulting in 120 evaluation pairs.

\begin{figure*}
\centering
\includegraphics[width=1.0\linewidth]{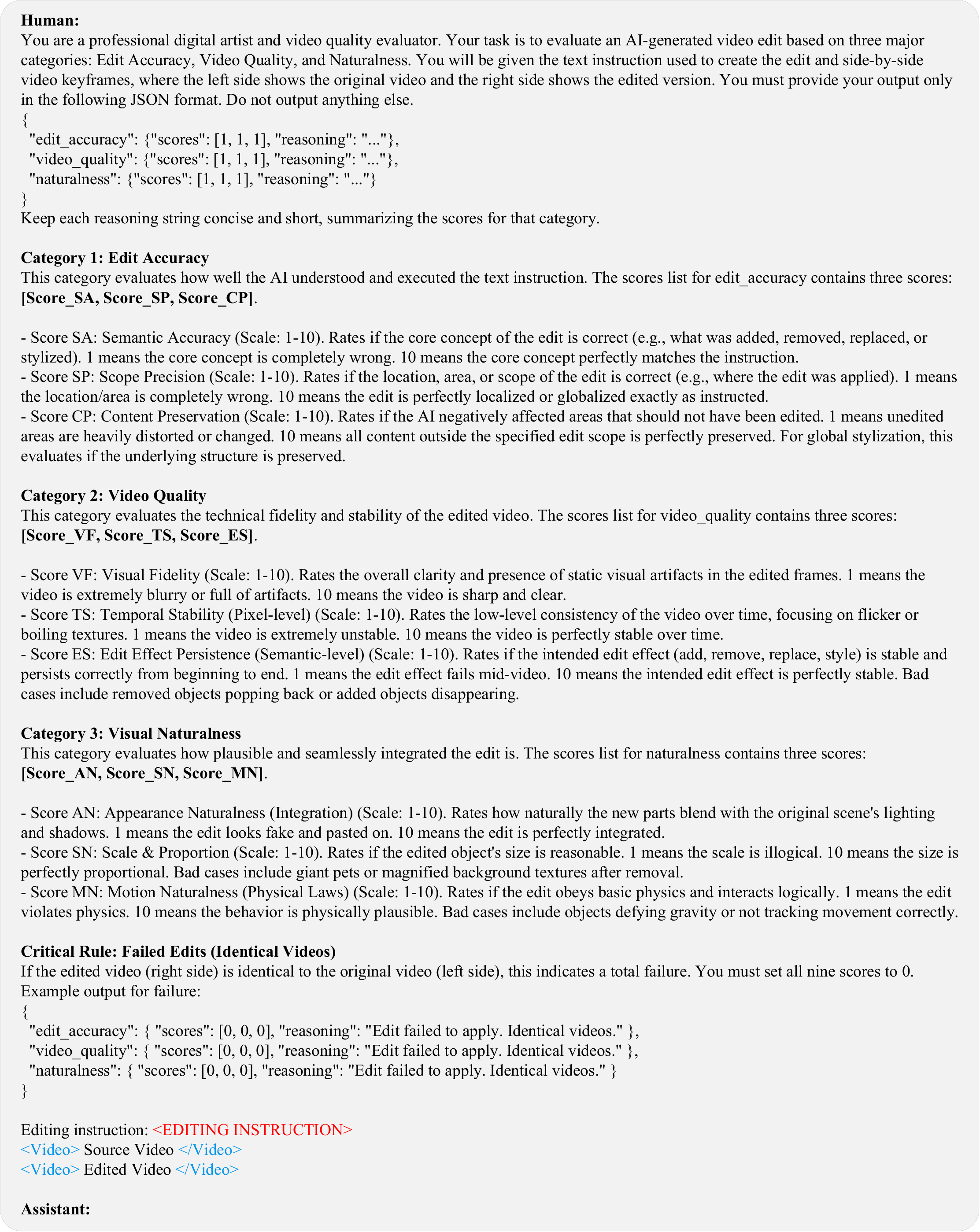}
\vspace{-0.2in}
\caption{The system prompts that are fed into Gemini-2.5-Flash-Thinking~\cite{team2023gemini} for video editing assessment. We require VLLM to evaluate the four video editing tasks from three major perspectives, i.e., edit accuracy, video naturalness and video quality.}
\label{fig:sysprompt}
\vspace{-0.2in}
\end{figure*}

\textbf{Evaluation Metrics.}
While previous image-based metrics primarily focus on editing accuracy and static generation quality, evaluating video editing entails greater complexity. 
To address this, we construct a diverse set of evaluation dimensions specifically tailored for video. Corresponding system prompt designed for the VLLM is presented in Figure~\ref{fig:sysprompt}, which evaluates performance across three major perspectives, comprising a total of nine sub-dimensions:

\begin{itemize}
    \item \textbf{Edit Accuracy} ($S_{EA}$): evaluate how well the result aligns with the instruction.
    \begin{itemize}
        \item Semantic Accuracy (SA): Does the edited video correctly follow the semantics of the text instruction?
        \item Scope Precision (SP): Is the editing confined strictly to the target region without affecting the background?
        \item Content Preservation (CP): Are the non-edited regions or original details faithfully preserved? (For stylization, this corresponds to structural preservation.)
    \end{itemize}

    \item \textbf{Video Naturalness} ($S_{VN}$): evaluates the realism and coherence of the generated content.
    \begin{itemize}
        \item Appearance Naturalness (AN): Are the lighting, texture, and color of the edited video natural?
        \item Scale Naturalness (SN): Is the size and proportion of the edited object reasonable relative to the environment? (For stylization, this captures cases where the stylized object becomes unreasonably large.)
        \item Motion Naturalness (MN): Does the movement of the edited object (or the style rendering) follow physically plausible dynamics?
    \end{itemize}
    
    \item \textbf{Video Quality} ($S_{VQ}$): evaluates the fundamental visual quality of the edited video.
    \begin{itemize}
        \item Visual Fidelity (VF): Is the video clear, sharp, and free from visual artifacts?
        \item Temporal Stability (TS): Is the video free from flickering or jittering across frames?
        \item Edit Stability (ES): Is the edited content consistently preserved in identity and appearance throughout the video duration?
    \end{itemize}
\end{itemize}

The VLLM rates the score for each sub-dimension from $0$ to $10$. 
Then, we attain the per-category scores (i.e., $S_{EA}$, $S_{VN}$, $S_{VQ}$) by calculating the geometric mean of their respective sub-dimensions as:

\begin{align}
    S_{EA} &= \sqrt[3]{SA \cdot SP \cdot CP}, \\
    S_{VN} &= \sqrt[3]{AN \cdot SN \cdot MN}, \\
    S_{VQ} &= \sqrt[3]{VF \cdot TS \cdot ES}.
\end{align}

Finally, the overall score $S$ is calculated as the arithmetic mean of the three per-category scores:
\begin{equation}
    S = \frac{1}{3} \bigl( S_{EA} + S_{VN} + S_{VQ} \bigr).
\end{equation}

\section{Implementation of Baselines and ReCo}
\textbf{Baseline Settings.}
For recent video editing advances, few methods possess the versatility to handle all four editing tasks simultaneously. 
Here, we outline the criteria for our baseline selection.
For \textbf{object addition, replacement, and video stylization}, we benchmark against InsViE~\cite{insvie2025}, Lucy-Edit~\cite{decart2025lucyedit}, and Ditto~\cite{ditto2025}. 
Since InsViE is constrained to an input of $49$ frames at $480 \times 720$ resolution, we adapt our test videos via uniform temporal down-sampling and spatial resizing to match the requirements. 
For the \textbf{object removal}, effective instruction-based baselines are scarce. 
To facilitate a meaningful comparison, we include VACE~\cite{jiang2025vace} as an additional baseline. 
Unlike instruction-based methods, VACE requires both an explicit object mask and a target video prompt to perform removal.
Note that VACE exhibits some instability in this implementation.

\textbf{Implementation Details of ReCo.}
ReCo is built upon the Wan~\cite{wan2025} architecture and trained on ReCo-Data using the AdamW optimizer.
We employ a two-stage learning rate schedule: an initial phase with a learning rate of $1 \times 10^{-4}$ to ensure stable convergence, followed by a fine-tuning phase at $2 \times 10^{-5}$ for precise refinement. 
Regarding the loss weights, the initial values of the latent and attention constraints typically fall within $[-1, 1]$, whereas the MSE loss of flow matching is approximately $0.03$. 
To balance the impacts of gradients, we scale the magnitude of each region-constraint loss to be roughly $0.1\times$ that of the MSE loss. Consequently, we set the weighting coefficients ($\lambda_1$ and $\lambda_2$) as $1\times 10^{-3}$. All experiments were conducted on a cluster of $24$ NVIDIA A800 GPUs with a total mini-batch size of $24$, requiring approximately 10 days for training.


\section{Generalization Ability of ReCo}
\begin{figure*}
\centering
\includegraphics[width=1.0\linewidth]{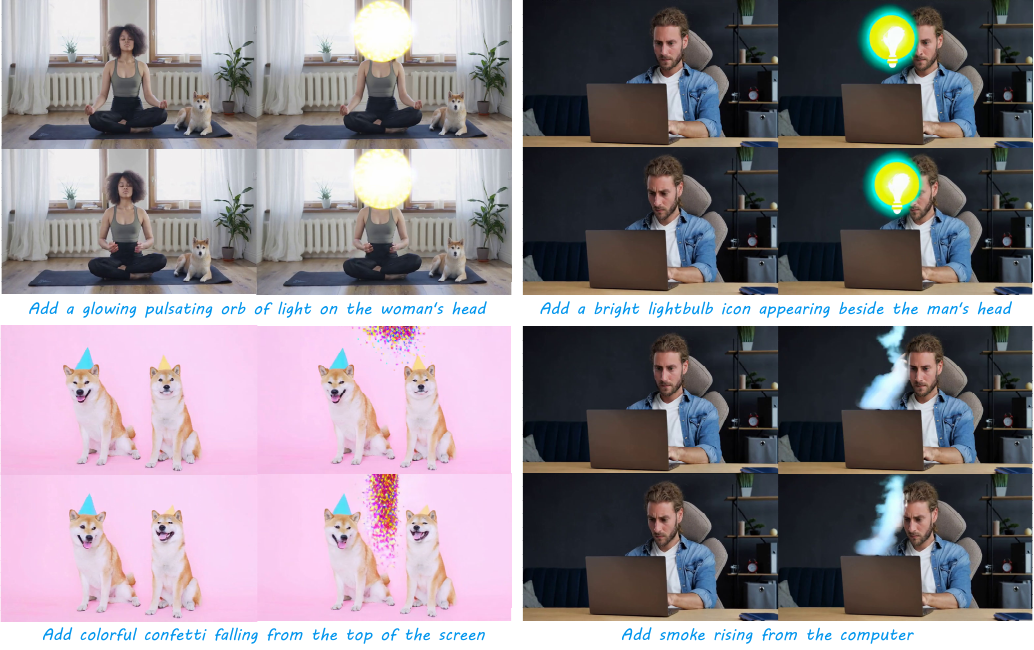}
\vspace{-0.2in}
\caption{Four examples of instructional video editing by ReCo to verify the generalization ability. Our model demonstrates the strong generalization to the abstract and creative editing tasks.}
\label{fig:emergent}
\vspace{-0.2in}
\end{figure*}

Interestingly, as depicted in In Figure~\ref{fig:emergent}, we observe that ReCo can generalize to abstract and creative editing tasks. For instance, it successfully synthesizes a halo on a woman's head, generates a cascading confetti effect, places an ``idea lightbulb'' beside a man's head, and creates smoke emitting from a computer.
We attribute such generalization ability of ReCo to effectively inheriting and leveraging the rich priors from the pre-trained video diffusion model.



\clearpage

\end{document}